\documentclass[10pt, twoside, twocolumn]{IEEEtran}

\usepackage{times}
\usepackage{epsfig}
\usepackage{graphicx}
\usepackage{amsmath}
\usepackage{amssymb}
\usepackage{helvet}
\usepackage{courier}
\usepackage{graphicx}
\usepackage{subfigure}
\usepackage{color}
\usepackage{amssymb}
\usepackage{algorithm}
\usepackage{algorithmic}
\usepackage{multirow}
\usepackage{xcolor}
\usepackage{multirow}
\usepackage{longtable}
\usepackage{url}

\begin{document}

\title{Stabilizing Training of Generative Adversarial Nets \\via Langevin Stein Variational Gradient Descent}

\author{Dong Wang, Xiaoqian Qin, Fengyi Song and Li Cheng \thanks{Dong Wang is a postdoctoral research fellow with the Bioinformatics Institute, A*STAR, Singapore.
   Xiaoqian Qin is an associate professor in the School of Urban and Environmental Sciences at Huaiyin Normal University, China. Fengyi Song is an associate professor in the School of Computer Science and Technology at Nanjing Normal University, China. Li Cheng is an associate professor in the Department of Electrical and Computer Engineering, University of Alberta, Canada.
    Corresponding author: Dong Wang
    (dongwang@nuaa.edu.cn).}}


\maketitle

\maketitle

\begin{abstract}
Generative adversarial networks (GANs), famous for the capability of learning complex underlying data distribution, are however known to be tricky in the training process, which would probably result in mode collapse or performance deterioration. Current approaches of dealing with GANs' issues almost utilize some practical training techniques for the purpose of regularization, which on the other hand undermines the convergence and theoretical soundness of GAN. In this paper, we propose to stabilize GAN training via a novel particle-based variational inference --- Langevin Stein variational gradient descent (LSVGD), which not only inherits the flexibility and efficiency of original SVGD but aims to address its instability issues by incorporating an extra disturbance into the update dynamics. We further demonstrate that by properly adjusting the noise variance, LSVGD simulates a Langevin process whose stationary distribution is exactly the target distribution. We also show that LSVGD dynamics has an implicit regularization which is able to enhance particles' spread-out and diversity. At last we present an efficient way of applying particle-based variational inference on a general GAN training procedure no matter what loss function is adopted. Experimental results on one synthetic dataset and three popular benchmark datasets --- Cifar-10, Tiny-ImageNet and CelebA validate that LSVGD can remarkably improve the performance and stability of various GAN models.
\end{abstract}

\section{Introduction}

Recently deep generative models, especially generative adversarial networks (GANs)~\cite{goodfellow2014generative}, have achieved great successes in modeling complex, high dimensional data of images, speeches, and text, using deep neural nets and stochastic optimization. Take image analysis as an example, GANs have been explored in a wide range of problems including e.g. image synthesis~\cite{goodfellow2014generative} \cite{9042869}, image segmentation~\cite{wang2019thermal}, image-to-image translation~\cite{li2019asymmetric}, super-resolution~\cite{ledig2017photo}, feature embedding~\cite{hong2019gane}, segmentation~\cite{souly2017semi}, outlier detection \cite{liu2019generative}, denoising~\cite{chen2018image}, recommender systems~\cite{wang2019recsys}, zero-shot learning~\cite{gao2020zero}, which are mostly dealt within a unsupervised learning or semi-supervised learning paradigm.

The problem setting of a GAN typically involves producing realistic-looking images $x$ from a random input noise $z$, by employing a generator $G$ and a discriminator $D$ to engage the following minimax game formulation:
\begin{equation}\label{eq_gan}
\min_G \max_D E_{x\sim p}[\log D(x)] + E_{z\sim p_z}[\log(1-D(G(z)))]
\end{equation}
A typical training process would iteratively update the discriminator and generator to make sure the two nets are updated competitively and converge to a saddle point.

However, GANs training usually suffer from instability issues like mode collapse or vanishing gradient which probably result in performance degeneration. To address these issues, various kinds of approaches have been proposed. Basically, all these methods can be categorized into three types. The first type of methods is adopting practical training techniques to stabilize GAN training, such as \cite{radford2015unsupervised}, \cite{salimans2016improved} and \cite{jenni2019stabilizing}. The second type of methods is introducing explicit or implicit regularization into the training process. A typical one of them is Wasserstein GAN (WGAN) \cite{arjovsky2017wasserstein}, which can be seen as imposing a norm restriction on the model parameters. \cite{gulrajani2017improved} extends the weight clipping of WGAN to gradient penalty which further stabilizes the training process. \cite{li2018mr} employs a manifold regularizer for exploiting the geometry information of real data. \cite{miyato2018spectral} presents spectral normalization to regularize the discriminator. \cite{zhang2019self} introduces the attention mechanism to the GAN. Although the above methods, either adopting training tricks or introducing regularization, can empirically improve the stability of GAN training, they inevitably cause a deviation form the original min-max objective and thus potentially undermine the convergence and foundation of GAN.

The third type of methods is developing new modeling frameworks. These methods include $f$-GAN \cite{nowozin2016f} that proposes a generalized GAN formulation based on variational divergence minimization. Bayesian GAN \cite{saatci2017bayesian} that presents a Bayesian formulation of GAN which incorporates prior distributions to alleviate the mode collapse issue. Additionally, MCGAN \cite{mroueh2017mcgan} and Fisher-GAN \cite{mroueh2017fisher} utilize an novel integral probability metrics (IPM) framework. Although these methods provide us more options of GAN modeling, they in the meanwhile create new tricky optimization problems to tackle. More importantly, the mode collapse issue is still not addressed, as the main goal of generator is to cheat the discriminator rather than pursue diversity.

Notably, \cite{wang2016learning} gives a new insight into overcoming GANs' training issues, in which the author presented Stein-GAN model based on particle-based variational inference --- Stein variational gradient descent (SVGD) \cite{liu2016stein}. Different from MLE (maximum-likelihood) or MAP (maximum-a-poster) based estimation, SVGD imposes a repulsive force among particles for the purpose of matching the entire target distribution rather than only search a single optimum. Moveover, SVGD has a closed-form update, which promises us both flexibility and efficiency. However, there are still two main concerns of SVGD to be addressed. One is the particle degeneracy issue as discussed in \cite{zhuo2018message} which leads to underestimating the variance of target distribution and thus end up with similar result as mode collapse. The other issue exists in the convergence analysis of SVGD. Most of current studies such as \cite{liu2017stein, chen2018unified, zhang2018stochastic} interpret SVGD as a specific kind of Wasserstein gradient flow (WGF). However, this does not constantly hold as marked out in \cite{liu2019understanding} because SVGD restricts the update function to lying on a RKHS (reproducing kernel Hilbert space) which is not a well-defined Riemannian manifold. Therefore, SVGD should be just taken as an approximation to WGF rather than an exact WGF. These issues not only challenge the practicability of SVGD but also undermine its theoretical fundamentals.


To this end, we present a stable, efficient and theoretically sound particle-based variational inference --- Langevin SVGD (LSVGD). Our method still yields a closed-form particle-based update but differs from SVGD with an extra disturbance added. We demonstrate that by properly adjusting the noise variance, LSVGD simulates a Langevin process whose stationary distribution is exactly the target distribution. Furthermore, we show that LSVGD has an implicit regularization effect that encourage particles to escape from ``low-variance traps'', which boosts particles' spread-out and circumvents the particle degeneracy issue of SVGD. Additionally, we present an efficient way of applying particle-based variational inference (including LSVGD and SVGD) for training a typical GAN model regardless of the concrete loss function.


The remaining parts of this paper are organized as follows: In Sec.~\ref{sec_mov}, the background of SVGD is provided, then we present Langevin SVGD and conduct a thorough analysis in Sec.~\ref{sec_LSVGD}. In Sec.~\ref{sec_GAN}, we analyze the relationship between GAN training and Bayesian inference, and present how to apply LSVGD on training different types of GANs. In Sec.~\ref{sec_exp}, we demonstrate our method through a series of experiments on one synthetic dataset and three popular benchmark datasets, Cifar-10, Tiny-ImageNet and CelebA. We conclude this paper in Sec.~\ref{sec_conclusion}.

\section{Background and Motivation}\label{sec_mov}

\begin{figure*}[t!]
\center
\subfigure[]{\includegraphics[width=0.28\textwidth]{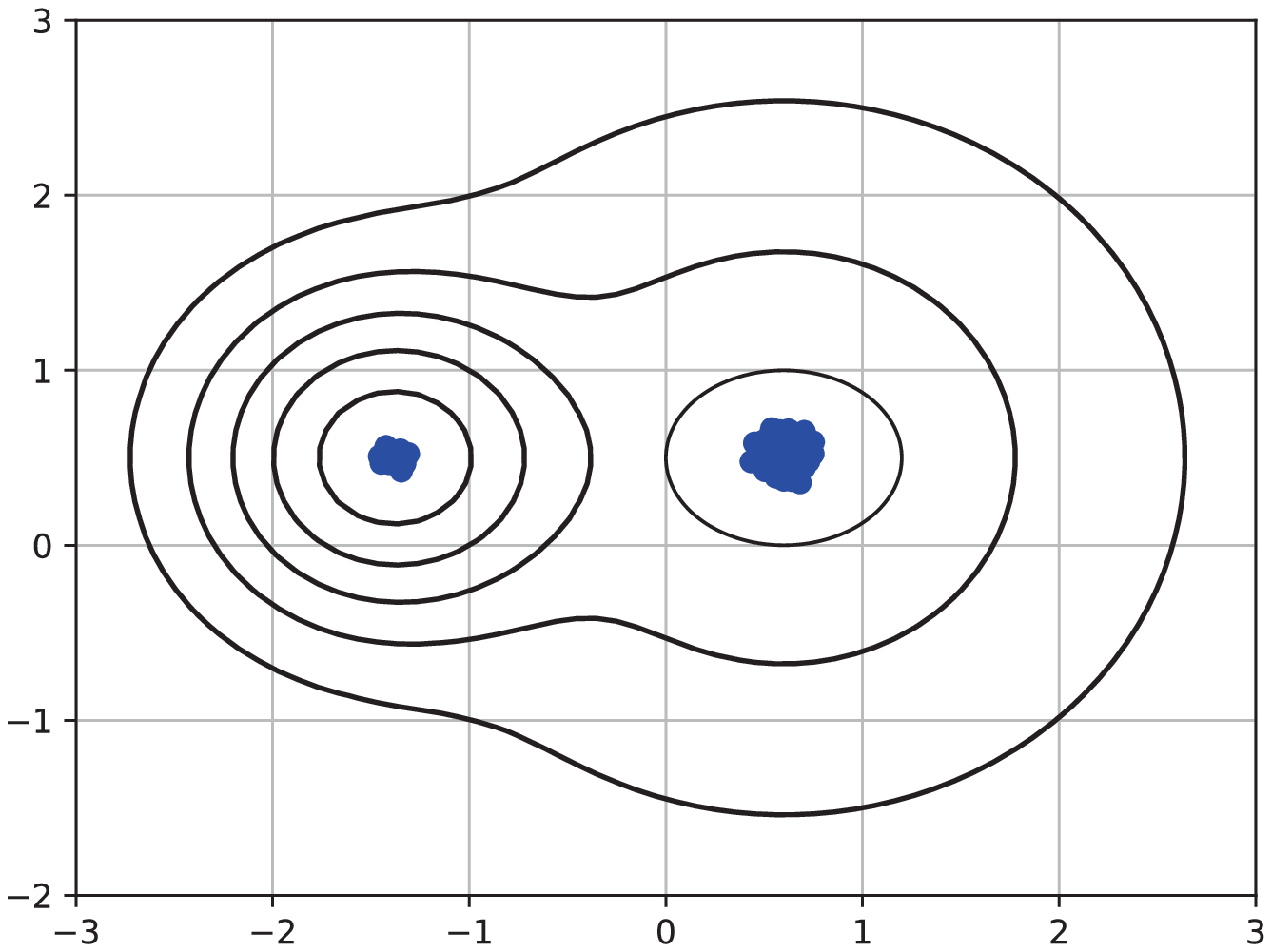}
\label{fig:toysvgd0}}
\subfigure[]{\includegraphics[width=0.28\textwidth]{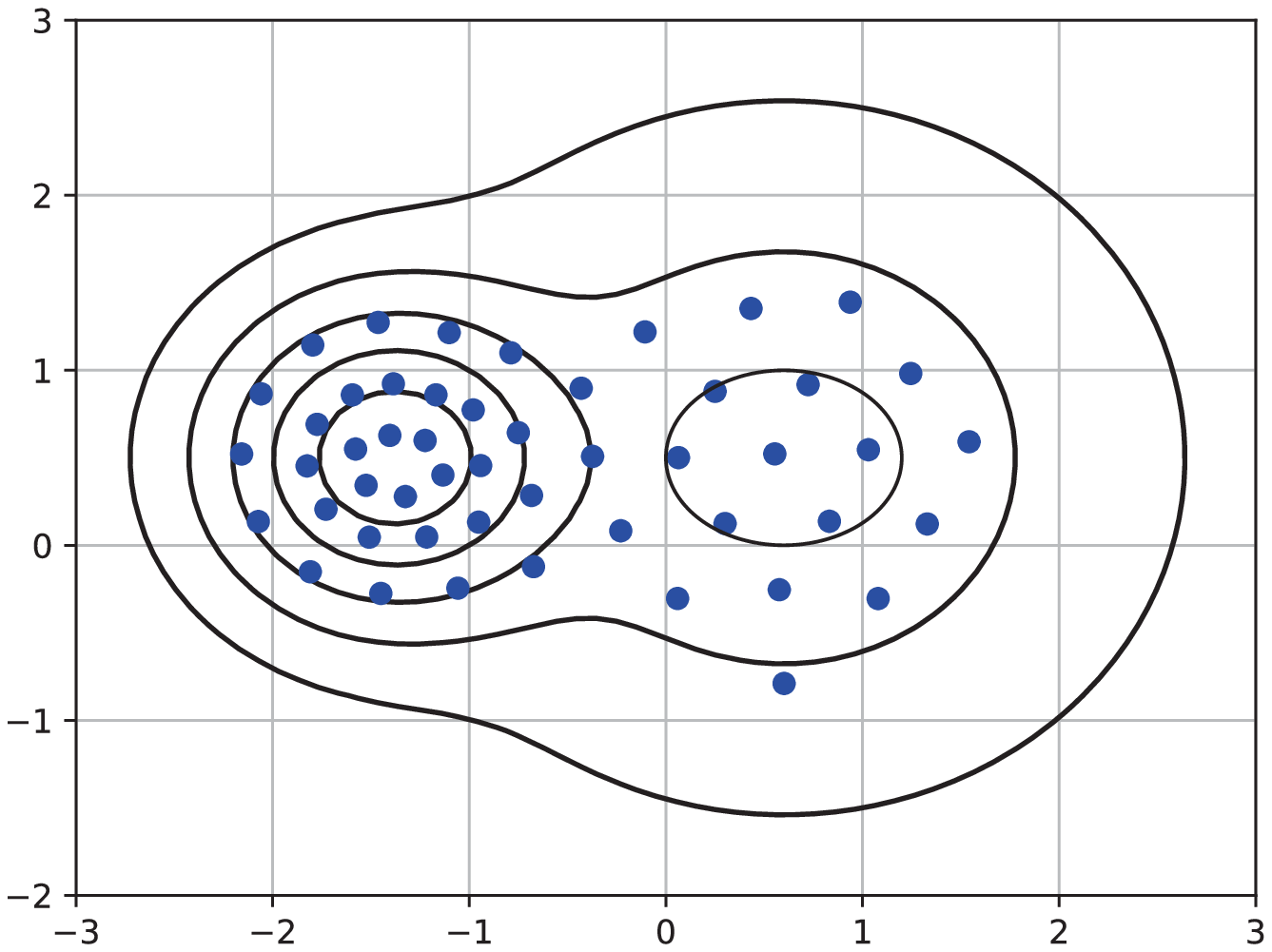}
\label{fig:toysvgd1}}
\subfigure[]{\includegraphics[width=0.28\textwidth]{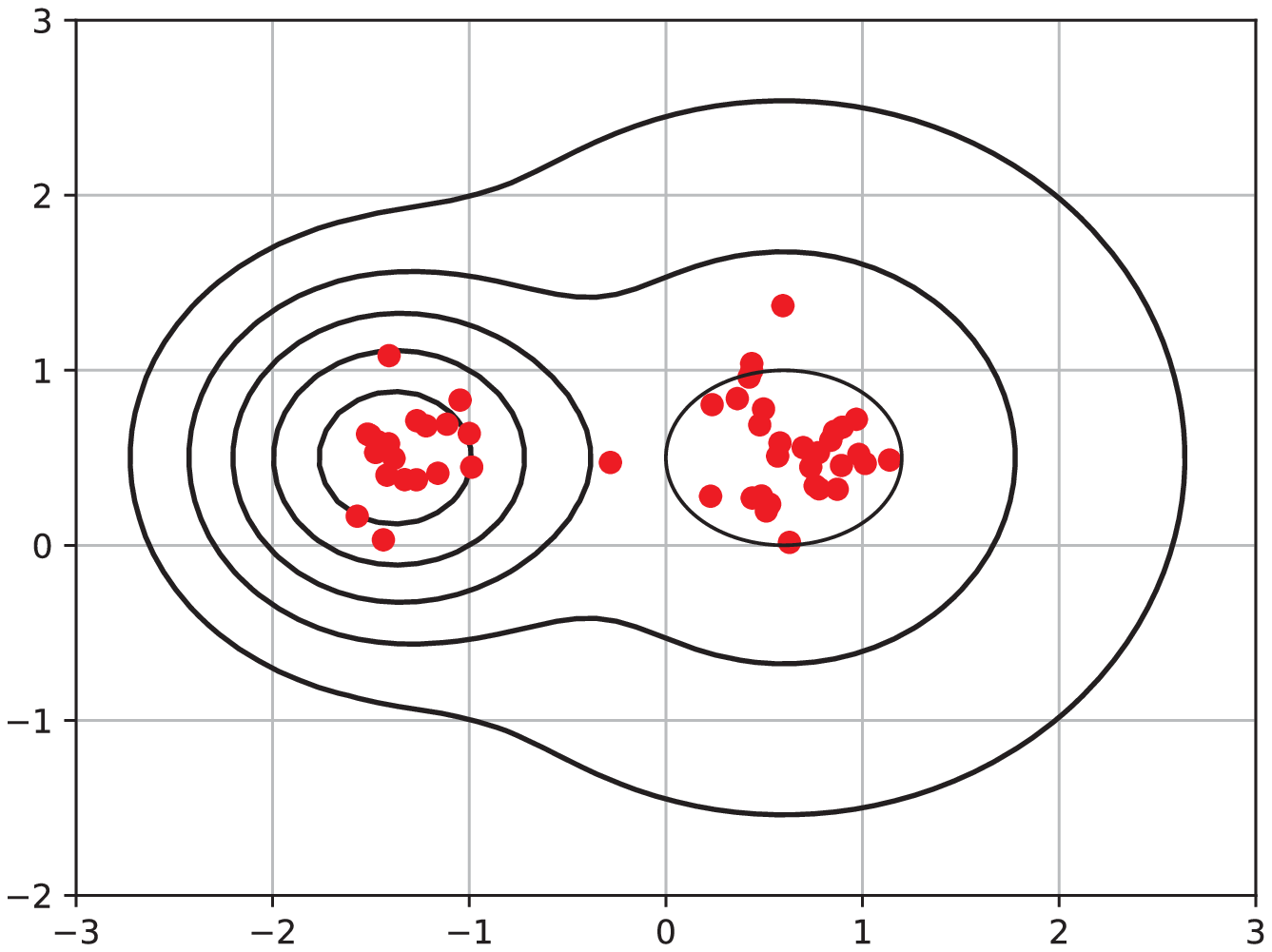}
\label{fig:toylsvgd}}
\caption{Comparison of SVGD and LSVGD. The target distribution is a bimodal Gaussian distribution over $\mathcal{R}^2$: $p(x) =$ $0.5\times\mathcal{N}(x|\mu_1, \sigma_1^2I)$ + $ 0.5\times \mathcal{N}(x|\mu_2, \sigma_2^2I)$ with $\mu_1=[-1,0.5]$, $\mu_1=[1,0.5]$, $\sigma_1^2 = 0.5$, $\sigma_2^2 = 1.0$ particles and the same initialization are used for both comparison methods. (a), (b) are the results (after 500 iterations) of SVGD with different kernel bandwidth: (a) $\gamma= 0.01$, (b) $\gamma= 0.1$. (c) is the result of LSVGD with $\gamma= 0.01$.}
\label{fig:toy}
\end{figure*}

Traditional variational Bayesian inference typically requires specifying a simple parametric approximate distribution, which potentially limits the feasible function space, and thus compromises inference quality. The Stein variational gradient descent (SVGD) method is introduced in~\cite{liu2016stein} to deal with this issue. In a nutshell, SVGD is a nonparametric variational inference method where the approximate distribution $q$ is defined as the empirical distribution of a set of particles $\{x_1,x_2,...x_n\}$. Hence, the complexity of this method depends on the number of particles used in the inference process. The goal of SVGD is to minimize the KL-divergence between the approximate distribution $q$ and the target distribution $p$. During training, each particle $x_i$ is updated with a small velocity field $\phi(x_i)$. Letting $x_i, \phi(x_i) \in \mathbb{R}^D$, then the updating rule is as follows:
\begin{equation}\label{eq_svgd}
T(x) = x + \delta \phi(x)
\end{equation}
where $\delta>0$ is a small step size.

Denote $S_p(x)$ the score function of $p$, i.e., $S_p(x) = \nabla_x \log p(x)$.
Let $\mathcal{A}_p$ be the Stein operator acting on a differentiable vector-valued function $\phi$, defined as
\begin{equation}
\mathcal{A}_p\phi(x) = S_p(x)\phi(x)^T + \nabla_x \phi(x)
\end{equation}
In this way, $\phi(x)$ behaves as a weight vector. It has been shown in \cite{liu2018stein} that by restricting the solution space $\mathcal{F}$ to lying on the unit ball of RKHS, then the steepest direction to decrease the KL-divergence between $q$ and $p$ is:
\begin{equation}\label{eq_steindis}
\phi^*(\cdot) = \mathrm{argmax}_{\phi \in \mathcal{F}} E_{x\sim q} \mathrm{tr}\left[\mathcal{A}_p\phi(x)\right]
\end{equation}
which has a closed-form solution:
\begin{equation}\label{eq_qphi}
\phi^*(\cdot) = E_{x\sim q}[\mathcal{A}_p k_{\gamma}(x,\cdot)] = \frac{1}{n}\sum_{i=1}^n[\mathcal{A}_p k_{\gamma}(x_i,\cdot)]
\end{equation}
where $\gamma$ is the corresponding kernel parameter. Specifically, we take the most widely used Gaussian kernel throughout this paper, i.e., $k_{\gamma}(x,y) = \exp(- \Arrowvert x_i - x_j\Arrowvert^2/\gamma)$ with $\gamma$ denoting the bandwidth in this case.

However, although SVGD has the particle efficiency and approximation flexibility compared with traditional inference methods, it also has one major issue within its current framework, that is the spread level of particles of SVGD mainly depends on the kernel parameter (e.g. the bandwidth $\gamma$). Therefore, the resulting match is fairly sensitive to the choice of kernel parameter and thus it is hard to capture the target distribution perfectly.

A toy example is illustrated in Fig.~\ref{fig:toysvgd0} and Fig.~\ref{fig:toysvgd1}. Here, two sets of resulting SVGD particles using different bandwidth are overlayed onto the target distribution $p$ that is defined as a bimodal Gaussian distribution: $p(x) = 0.5\times\mathcal{N}\left(x|\mu_1, \sigma_1^2I\right)$$~+~0.5\times \mathcal{N}\left(x|\mu_2, \sigma_2^2I\right)$, with $\mu_1=[-1,0]$, $\mu_1=[1,0]$, $\sigma_1^2 = 0.5$, and $\sigma_2^2 = 1.0$. When applying a small kernel bandwidth (e.g. 0.01) as in Fig.~\ref{fig:toysvgd0}, the repulsive force between particles is so weak that all particles concentrate around the MAP. On the contrary, when using a larger bandwidth, as in Fig.~\ref{fig:toysvgd1}, particles may be much deviated from their desired locations, which are also hard to capture the target distribution perfectly.

Owing to the difficulty in choosing parameter, to avoid particles being overly repelled in practice, a small $\gamma$ is often preferred, which consequently causes the particle degeneracy issue \cite{zhuo2018message} that accounts for the underestimation of the variance of target distribution. \cite{zhuo2018message} analysed that this issue could become pronounced if SVGD is performed in high dimensional space. To circumvent this issue, \cite{zhuo2018message} converts the original global inference problem of SVGD into a set of local low-dimensional ones. Nonetheless, this must highly raise the computational cost. Moreover, even in a low-dimensional space, SVGD still exhibits high sensitivity to the choice of kernel parameter (e.g., $\gamma$).

These investigations motivate us to develop a stable and efficient inference method for the approximate distribution $q$ to closely adhere to the high-density areas of target distribution $p$ while maintaining a sufficient level of spread-out with lower parameter sensitivity.

\section{Langevin Stein Variational Gradient Descent}\label{sec_LSVGD}
In this section, we first propose Langevin SVGD (LSVGD) and give an in-depth analysis of its properties (in Theorem 1 and Theorem 2), then present an effective way to apply LSVGD on various types of GANs.

The basic idea of LSVGD is replacing the update of SVGD (as Eq.~\eqref{eq_svgd}) with the following noise injected update:
\begin{equation}\label{eq_LSVGD}
\begin{split}
T_\epsilon(x) &= T(x+\epsilon(x))\\
 &= x+ \epsilon(x) + \delta \phi(x+ \epsilon(x))
\end{split}
\end{equation}
where $\epsilon(x)$ denotes the random noise vector sampled from a zero-centered Gaussian distribution $\mathcal{N}(0,\Sigma(x))$ with covariance matrix entirely depending on $x$. Notably, each particle $x_i$ takes with an independent noise $\epsilon(x_i)$. It has already been advocated in a prior empirical study \cite{neelakantan2015adding} that adding noise to gradient descent method facilitates the training process to escape unstable saddle points. In SVGD literature, this mechanism works as a complement to the kernel-based repulsive force, which significantly reduces the sensitivity to the parameter setting as shown in Fig.~\ref{fig:toylsvgd}.

Without loss of generality, we start our analysis from a general covariance $\Sigma(x)$, and then detail how to adapt it in Theorem.~1. In the following Lemma, we first show the relationship between LSVGD and standard SVGD, based on which we create an efficient updating rule for LSVGD.

\textbf{Lemma~1}: For a set of $n$ particles $\{x_1,x_2,...,x_n\}$, sample from $\epsilon(x)\sim \mathcal{N}(0,\Sigma(x))$ the corresponding noise vectors. Let $T_\epsilon(x)$ be a bijective and differentiable function for any specific sample $\epsilon$, and $q_{[T_\epsilon]}$ the distribution of particles transformed with $T_\epsilon(x)$, then we have,
\begin{equation}\label{eq_lemma1}
E_{\epsilon\sim \mathcal{N}(0,\Sigma(x))} \frac{\partial}{\partial\delta} \mathrm{KL}\left(q_{[T_\epsilon]}||p\right) = -E_{x\sim q_\epsilon}\mathrm{tr} \left[\mathcal{A}_{p}\phi(x)\right]
\end{equation}
where $q_\epsilon$ a $n$-component Gaussian mixture distribution with each component centered at $x_i$ with covariance $\Sigma(x_i)$.

From Eq.~\eqref{eq_lemma1}, we define LSVGD's update as the steepest descent that decreases the expected noise perturbed KL-divergence. Note that the only difference between LSVGD and SVGD is that in each iteration we replace the discrete distribution $q$ with its continuous counterpart $q_\epsilon$ (with the noise level property set). Similar to SVGD, by restricting the solution to lying on the unit ball of RKHS, we have the following closed-form solution:
\begin{equation}\label{eq_qephi}
\phi^*(\cdot) = E_{x\sim q_{\epsilon}}[\mathcal{A}_p k_{\gamma}(x,\cdot)]
\end{equation}
Since $q_{\epsilon}$ is a continuous distribution, we adopt the following importance sampling approximation to boost efficiency:
\begin{equation}\label{eq_imqephi}
\phi^*(\cdot) = \sum_{i=1}^n w_i \cdot \mathcal{A}_p k_{\gamma}(x_i,\cdot)~,~w_i = \frac{q_{\epsilon}(x_i)}{\sum_{j=1}^nq_{\epsilon}(x_j)}
\end{equation}
Finally, by plugging Eq.~\eqref{eq_imqephi} back into Eq.~\eqref{eq_LSVGD}, we obtain the final updating rule of LSVGD:
\begin{equation}\label{eq_LSVGDfm}
\begin{split}
T_\epsilon(x) =&~x + \delta \sum_{i=1}^n w_i \cdot [k_{\gamma}(x_i,x)\nabla_{x_i} \log p(x_i)\\
&+\nabla_{x_i}k_{\gamma}(x_i,x)]+ \epsilon(x)
\end{split}
\end{equation}
The above analysis shows that LSVGD inherits the particle efficiency of SVGD while incorporating extra randomness into its training process. In the following theorem, we will show how to properly adjust the noise magnitude with which particles will eventually converge to the target distribution.

\textbf{Theorem~1}: By properly adjusting the injected noise $\epsilon(x)\sim\mathcal{N}(0,\Sigma(x))$, particles evolving with LSVGD (eq.~\eqref{eq_LSVGDfm}) follow a Langevin process whose stationary distribution is exactly the target distribution $p(x)$.

We give a sketch of proof here, and please refer to appendix for a detailed proof. Our proof is built on the following assumptions: First, for simplicity, we regard $w_i$ as constants during each iteration. Second, we assume $\gamma$ defines an active neighborhood $\Omega_x^R$ (with radius $R$) for each particle $x$, i.e., truncating $k(x_i,x)$ to 0 for all $x_i$ with $\Arrowvert x_i-x \Arrowvert_2>R$. This is because the magnitude of Gaussian kernel $k_{\gamma}(x_i, x)$ decreases exponentially as $x_i$ getting away from $x$. Third, we assume that for any $x_i$ inside $\Omega_x^R$, the gradient $\nabla_{x_i}\log p(x_i)$ is a random variable following a Gaussian distribution: $\nabla_{x_i}\log p(x_i) = \nabla_{x}\log p(x) + \mathcal{N}(0,V(x))$.

Based on the above assumptions, Eq.~\eqref{eq_LSVGDfm} is equivalent to:
\begin{equation}
T_\epsilon(x)-x = \delta [B(x)\nabla_{x}\log p(x)-\Gamma(x)] +\mathcal{N}(0,\Sigma(x))
\end{equation}
which is a special case of the full recipe of SGMCMC \cite{Ma2015SGMCMC} (by setting the curl matrix to 0). Here the drifting term $B(x)$ is defined as:
\begin{equation}
B(x)=\sum_{i=1}^n 1\{x_i \in \Omega_x^R\}\cdot w_i\cdot k_{\gamma}(x_i,x)\cdot I_D
\end{equation}
where $I_D$ denotes the $D\times D$ sized identity matrix. $\Gamma(x)$ is a $D$-dimensional vector with the $d$-th entry defined as:
\begin{equation}
\Gamma^d(x)= \sum_{l=1}^D \frac{\partial}{\partial x^l}(B^{d,l}(x))
\end{equation}
where $B^{d,l}(x)$ denotes entry ($d$,$l$) of $B(x)$.
where $x^l$ denotes the $l$-th dimension of $x$. Then according to \cite{Ma2015SGMCMC}, by adjusting the covariance matrix $\Sigma(x)$ as:
\begin{equation}
\Sigma(x)=\delta(2B(x)-\delta V(x))
\end{equation}
Eq.~\eqref{eq_LSVGDfm} can be seen as a discretization of the stochastic differential equation whose stationary distribution is exactly $p(x)$. This explains the convergence behavior of LSVGD as shown in Fig.~\ref{fig:toylsvgd}. In the implementation, since $V(x)$ is unknown, we use its empirical estimation $\hat{V}(x)$ instead\footnote{We choose a threshold $0.001$, and all $k_\gamma(x_i,x)$ below this threshold is set to 0. Thus, each $\hat{V}(x)$ is calculated with those valid $\nabla_{x_i}\log p(x_i)$ with $k_\gamma(x_i,x)>0$.}.
Note that dealing with LSVGD in the full recipe of SGMCMC is more theoretically sound than interpreting it as Wasserstein gradient flow, due to the fact that SVGD is performed in RKHS which is not a well-defined Riemannian manifold.


In the following theorem, we analyse the regularization effect of LSVGD which helps to circumvent the particle degeneracy issue.

\textbf{Theorem~2}: Assume that both $\phi(x)$ and $S_p(x)$ can be well approximated with the first-order Taylor expansions by considering a sufficiently small noise $\epsilon$.
Denote $H_p(x)$ the Hessian matrix of $\log p(x)$ at $x$, and $R(q, p) := -E_\epsilon E_{x\sim q}\epsilon^T \nabla_x\phi(x) H_p(x)\epsilon$ a quadratic term of $\epsilon$.
Then the expected gradient of the KL-divergence of the noise injected update can be decomposed into two terms,
\begin{equation}\label{eq_gradkl}
E_{\epsilon}\frac{\partial}{\partial\delta} \mathrm{KL}(q_{[T_\epsilon]}||p) = \frac{\partial}{\partial\delta}\mathrm{KL}(q_{[T]}||p) + R(q, p)
\end{equation}

We give an overall analysis here. Please refer to appendix for more details. $E_{\epsilon}\frac{\partial}{\partial\delta} \mathrm{KL}(q_{[T_\epsilon]}||p)$ quantifies how $q$ matches its target $p$, which gets reduced gradually as $q$ converges to $p$. This indicates that the magnitude of both $\frac{\partial}{\partial\delta}\mathrm{KL}(q_{[T]}||p)$ and $R(q, p)$ would decrease as the training process going towards a stable optimum. Therefore, alongside minimizing the KL-divergence, LSVGD also treats an implicit regularization term. Let's take an instance a $K$-component Gaussian mixture model: $p(x) = \sum_{k=1}^K \pi_k \mathcal{N}(x| \mu_k, \sigma_k^2I)$, with $x, \phi(x)$ $\in R^d$, and let $\nabla_x\phi(x) = -C_{\phi}I$ for simplicity, then both $\frac{\partial}{\partial\delta}\mathrm{KL}(q_{[T]}||p)$ and $R(q, p)$ are negative. Then, we have
\begin{equation}
|R(q, p)|\approx d\sigma^2 C_{\phi}\sum_k \sigma_k^{-2}E_{x\sim q}1\{x \in \Omega_k\}
\end{equation}
with $\Omega_k$ indicating the $k$-th component. We can see that $R(q, p)$ acts as a regularization that encourages $q$ to place its probability mass on areas with higher variance ($\sigma_k^{2}$). In this way, it facilitates particles to escape from ``low-variance traps", which consequently enhances particles' spread-out and circumvents the particle degeneracy issue. Additionally, this mechanism also works as a complement to the kernel-based repulsive force, which significantly reduces its sensitivity to the parameter setting as shown in Fig.~\ref{fig:toylsvgd}.

\section{Applying LSVGD on GAN Training}\label{sec_GAN}
In this section, we first analyse the relationship between GAN Training and Bayesian Inference, then present how to introduce LSVGD to GAN training. We further extend our method for training conditional GANs with auxiliary classifier.

\subsection{Relationship between GAN Training and Bayesian Inference}
As first analysed in \cite{goodfellow2014generative}, under some assumptions, the original GAN training is equivalent to minimizing the Jensen-Shannon divergence (JSD) between real and generated data distributions. However, in a practical training process, generator and discriminator are almost trained in a balanced way of competing against each other for yielding valid gradient, which violates the basic assumption of \cite{goodfellow2014generative} that discriminator has infinite capacity. Besides that, f-GAN \cite{nowozin2016f} also tries to build a link between GAN training and Bayesian inference, which defines a family of variational GAN models based on variational divergence minimization. However, this modeling is only workable for limited kinds of GANs within the f-GAN family. In this work, we present a natural and simple Bayesian interpretation for a common GAN training by exploiting particle-based variational inference.

For a common GAN architecture, the discriminator $D(x)$ can be seen as concatenation of a feature extractor $f_{D}(x)$ \footnote{In a classification network, the output of the second outermost layer is usually used as features for input data. Therefore, a simple division is that all the layers before the outermost layer are regarded as feature extractor, while the remaining part is regarded as classifier.} and a classifier $l_{D}(f_{D}(x),y)$. Denote $y$ and $y_z$ respectively label of $x$ and $G(z)$. Denote $\{f_{D}(x_i)\}_{i=1}^n$ and $\{f_{D}(G(z_i))\}_{i=1}^n$ a training batch sampled from real and generated data respectively. The objective for training generator is:
\begin{equation}\label{eq_obG}
\begin{split}
J_G :=& \min_{G} \frac{1}{n}\sum_{i=1}^n l_{D}(f_{D}(G(z_i)),y_z=+1)
\end{split}
\end{equation}
By defining a likelihood function:
\begin{equation}\label{eq_pobj}
p(y|f_{D}(x)) \propto \exp(-l_{D}(f_{D}(x),y))
\end{equation}
Then, minimizing $J_G$ follows a standard MLE routine that iteratively raises the likelihood of generated data $\{f_{D}(G(z_i))\}_{i=1}^n$ with label being set to $+1$. Denote $\hat{q}$ the empirical distribution of $\{f_{D}(G(z_i))\}_{i=1}^n$. Then Eq.~\eqref{eq_obG} can be seen as an approximation to the following Bayesian inference task
\begin{equation}\label{eq_obGKL}
J_G := \min_{G} \mathrm{KL}(\hat{q}||p(y_z=+1|f_{D}(G(z))))
\end{equation}
where the entropy term of $\hat{q}$ disappears. The original objective of discriminator is defined as:
\begin{equation}\label{eq_obD}
\begin{split}
J_D := \min_{D}~&\frac{1}{n}\sum_{i=1}^n l_{D}(f_{D}(x_i),y=+1) \\
&+\frac{1}{n}\sum_{i=1}^n l_{D}(f_{D}(G(z_i)),y_{z_i}=-1)
\end{split}
\end{equation}
Similarly, minimizing $J_D$ is to raise the likelihood of both real and generated data with label flipped to $-1$. Denote $\hat{p}$ the empirical distribution of $\{f_{D}(x_i)\}_{i=1}^n$. Then Eq.~\eqref{eq_obD} shares the same goal with the following objective:
\begin{equation}\label{eq_obDKL}
\begin{split}
J_D := \min_{D}~& \mathrm{KL}(\hat{p}||p(y=+1|f_D(x)))\\
& + \mathrm{KL}(\hat{q}||p(y_z=-1|f_D(G(z))))
\end{split}
\end{equation}
In this way, both generator and discriminator training can be dealt within divergence minimization framework. Hence, GAN training can be seen as performing Bayesian inference for a moving target whose distribution varies over time. In the following, we show that LSVGD can be applied on both generator and discriminator training.
\subsection{GAN Training with LSVGD}
In batch training, both $\{f_{D}(G(z_i))\}_{i=1}^n$ and $\{f_{D}(x_i)\}_{i=1}^n$ are used as particles to enhance diversity and address mode collapse issue of GAN. In detail, our approach is a two-stage procedure: First, the classification part keeps the normal way (SGD) of calculating gradient w.r.t. Eq.~\eqref{eq_obG} for generator and Eq.~\eqref{eq_obD} for discriminator respectively. Second, when gradient $\{\nabla f_{D}(G(z_i))\}_{i=1}^n$ or $\{\nabla f_{D}(x_i)\}_{i=1}^n$ arrives at the outermost layer of feature extractor $f_D(x)$, we replace it with the LSVGD version according to Eq.~\eqref{eq_LSVGDfm}, and back-propagate it to the remaining layers. This method not only provides a natural way of injecting extra disturbance into GAN training, but also shows how variational inference can be applied on neural network.


\subsection{Conditional GAN with Auxiliary Classifier}
\begin{figure}[t!]
\centering
\includegraphics[width=0.49\textwidth]{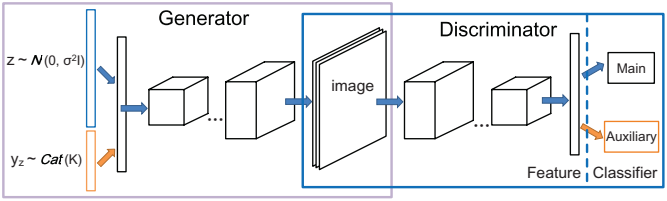}
\caption{Architecture of our conditional GAN.}
\label{fig:pipe}
\end{figure}
To enable supervised learning of GAN, we follow \cite{odena2017conditional} to utilize an augmented GAN architecture as shown in Fig.~\ref{fig:pipe}. Assume we have a labeled dataset $\{(x_i, y_i^C)\}_{i=1}^n$ with $y_i^C$ indicating class label. We feed the generator not only a noise vector $z$ but also a label vector $y_z^C$ which is uniformly sampled from the real label set $\{y_i^C\}_{i=1}^n$. We also add an auxiliary classifier $l_{D}^{C}(f_{D}(x),y^C)$ onto the discriminator which shares the same input features with the main classifier\footnote{The main classifier is used to classify an image being real or fake.} but aims to reconstruct the class labels.

For training the auxiliary classifier, we do not differentiate real $x$ and generated images $G(z,y_z^C)$, but map both of them to the corresponding labels. The loss function is defined straightforwardly:
\begin{equation}\label{eq_obAux}
\begin{split}
J_G^C &:= \min_{G} \frac{1}{n}\sum_{i=1}^n l_{D}^{C}(f_{D}(G(z_i,y_{z_i}^C)), y_{z_i}^C)\\
J_D^C &:= \min_{D} \frac{1}{n}\sum_{i=1}^n l_{D}^{C}(f_{D}(x_i), y_i^C)
\end{split}
\end{equation}
For applying LSVGD, we also adopt the following probabilistic version for Eq.~\eqref{eq_obAux} with likelihood function defined as $p^C(y^C|f_{D}(x)) \propto \exp(-l_{D}^{C}(f_{D}(x),y^C))$, which is
\begin{equation}\label{eq_obAuxKL}
\begin{split}
J_G^C &:= \min_{G} \mathrm{KL}(\hat{q}||p^C(y_z^C|f_{D}(G(z,y_z^C))))\\
J_D^C &:= \min_{D} \mathrm{KL}(\hat{p}||p^C(y^C|f_{D}(x)))
\end{split}
\end{equation}
Here the particles of $\hat{q}$ are augmented as $\{f_{D}(G(z_i,y_{z_i}^C))\}_{i=1}^n$ that includes class information.

\section{Empirical Experiments}\label{sec_exp}
We conduct extensive experiments to validate the performance of LSVGD. We first do experiments on a synthetic dataset to make a direct comparison of SVGD and LSVGD, then test the performance of applying LSVGD on various GANs on three popular benchmark datasets Cifar10 \cite{krizhevsky2009learning}, Tiny-ImageNet \cite{russakovsky2015imagenet} and CelebA \cite{liu2015faceattributes}. 

\subsection{Experiments on Synthetic Data}
\begin{figure*}
\center
\subfigure[]{\includegraphics[width=0.28\textwidth]{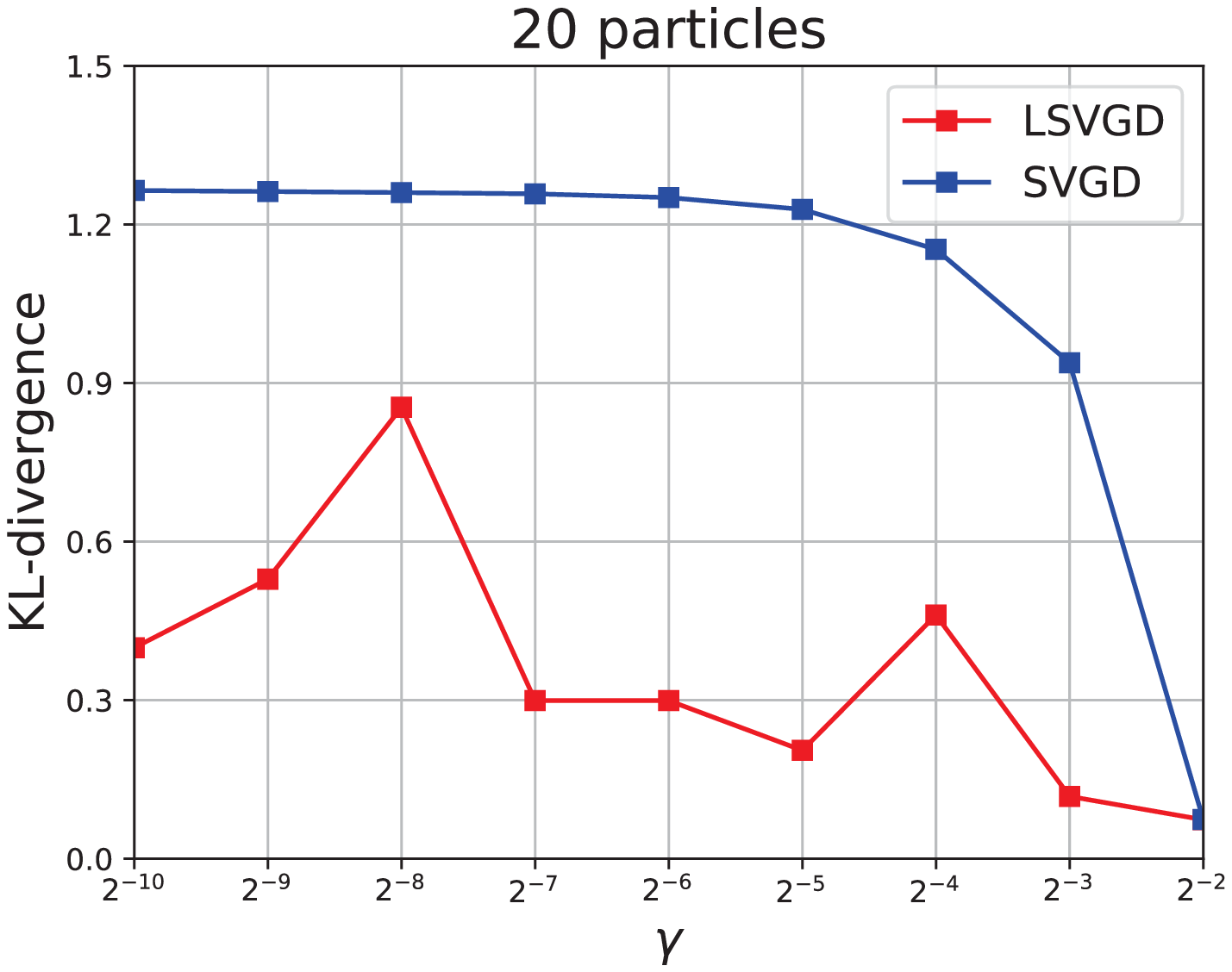}
\label{fig:kl20}}
\subfigure[]{\includegraphics[width=0.28\textwidth]{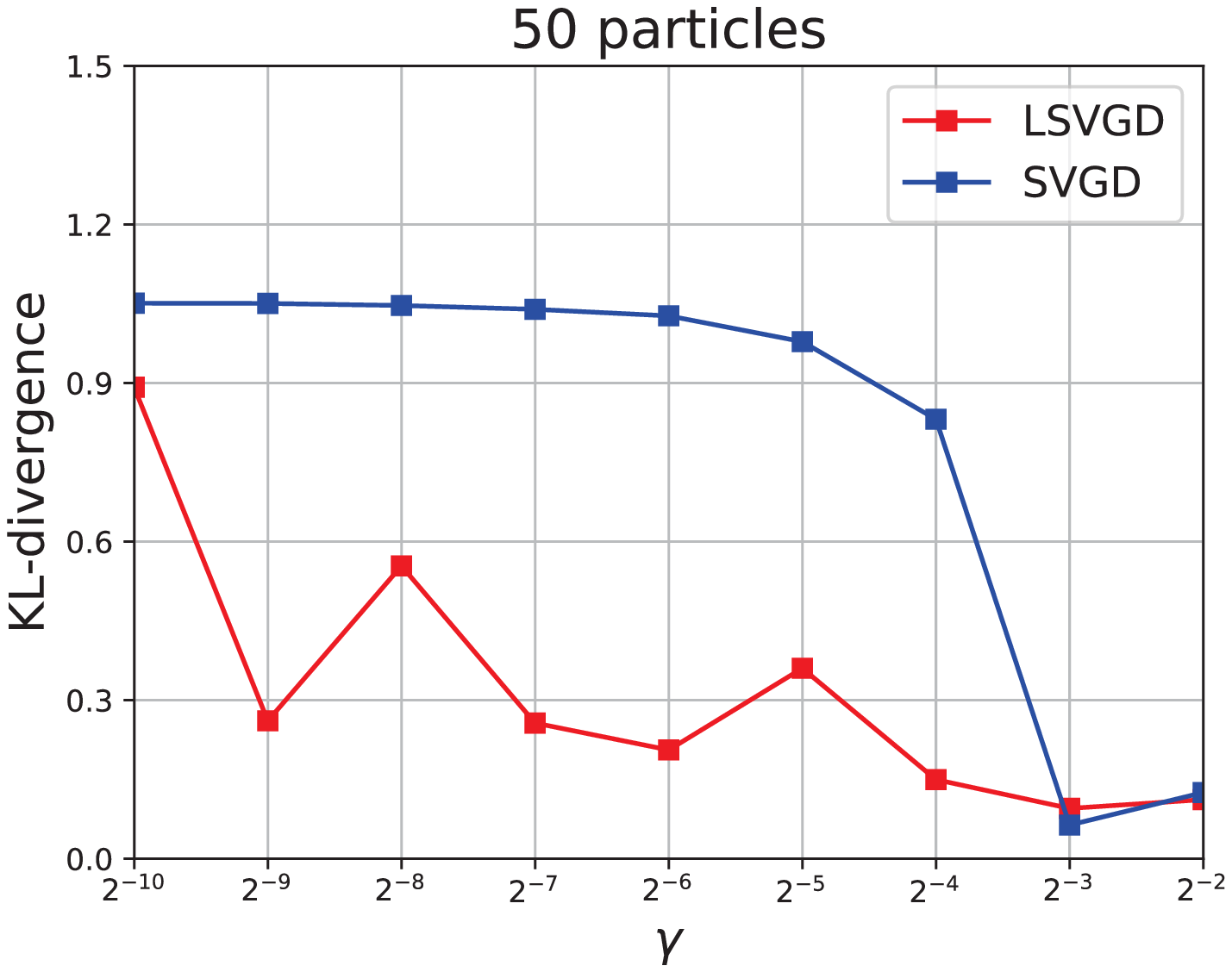}
\label{fig:kl50}}
\subfigure[]{\includegraphics[width=0.28\textwidth]{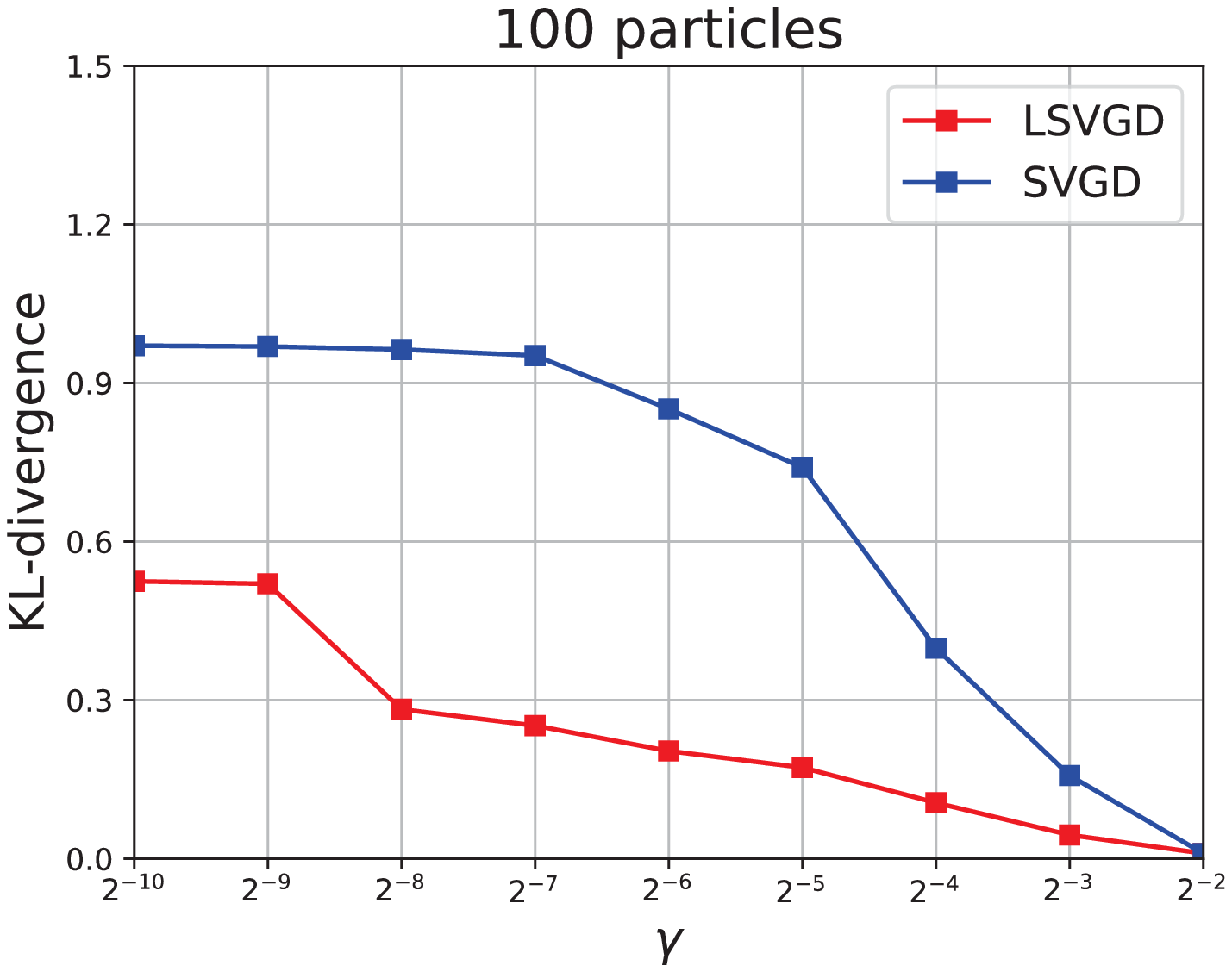}
\label{fig:kl100}}
\subfigure[]{\includegraphics[width=0.28\textwidth]{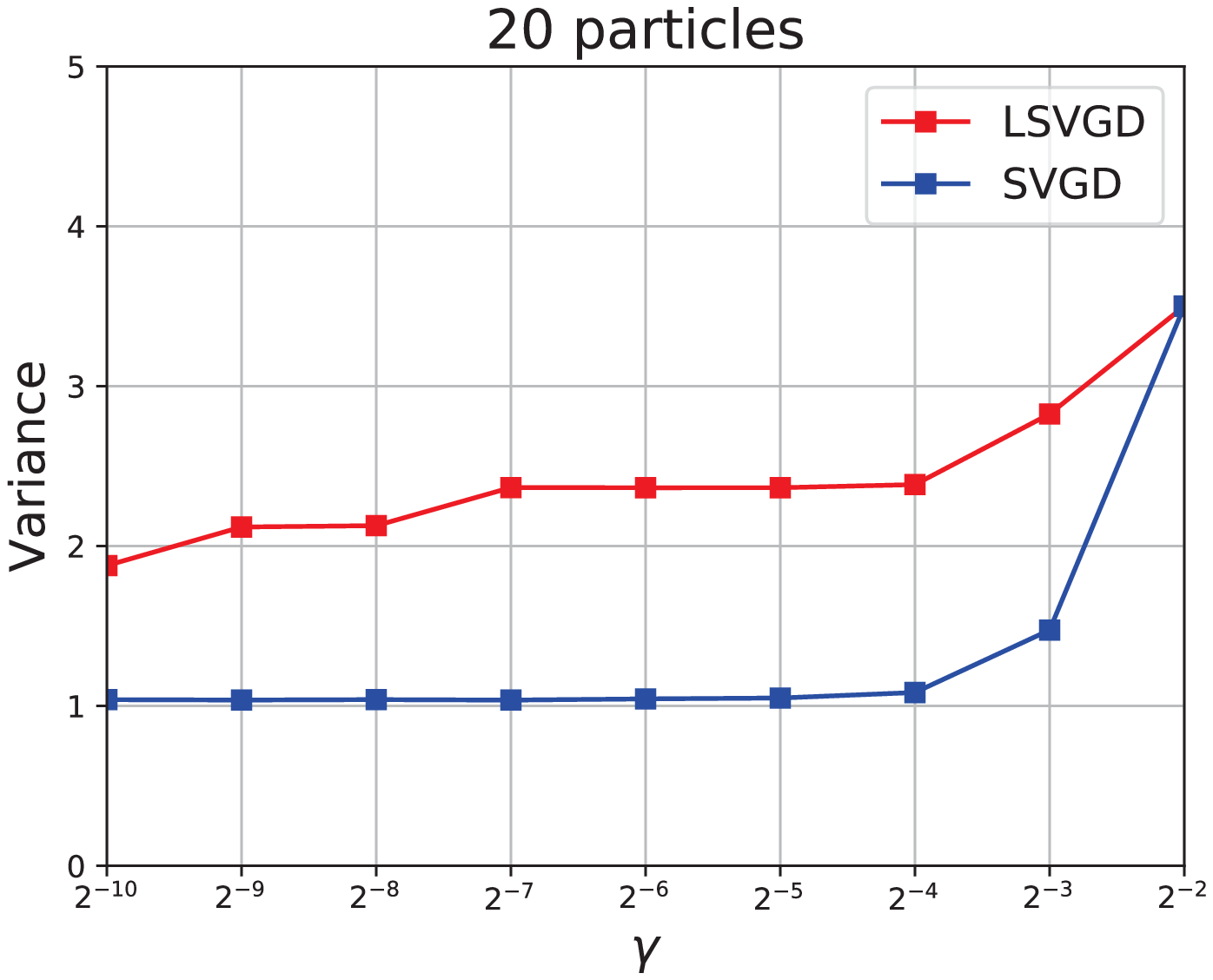}
\label{fig:var20}}
\subfigure[]{\includegraphics[width=0.28\textwidth]{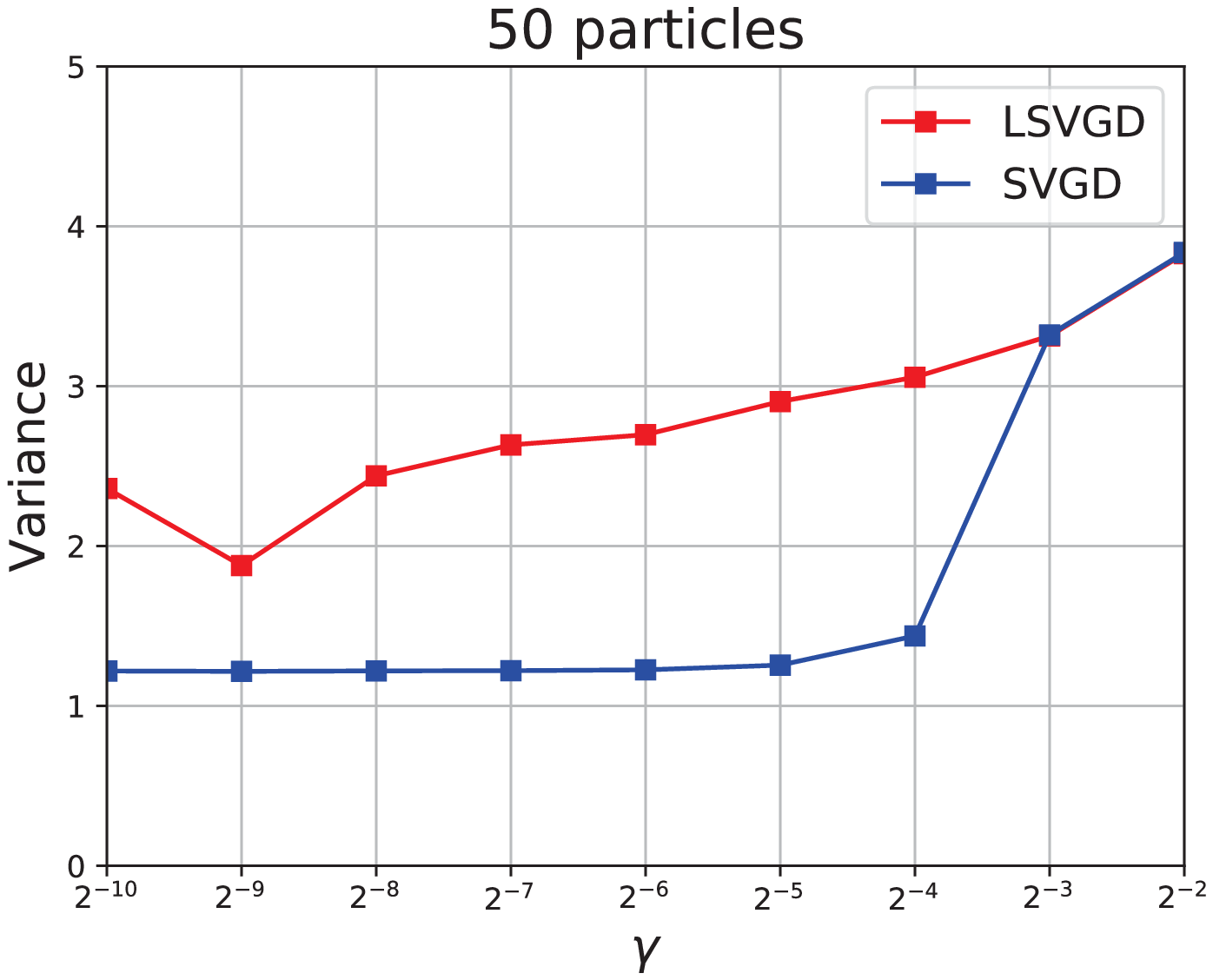}
\label{fig:var50}}
\subfigure[]{\includegraphics[width=0.28\textwidth]{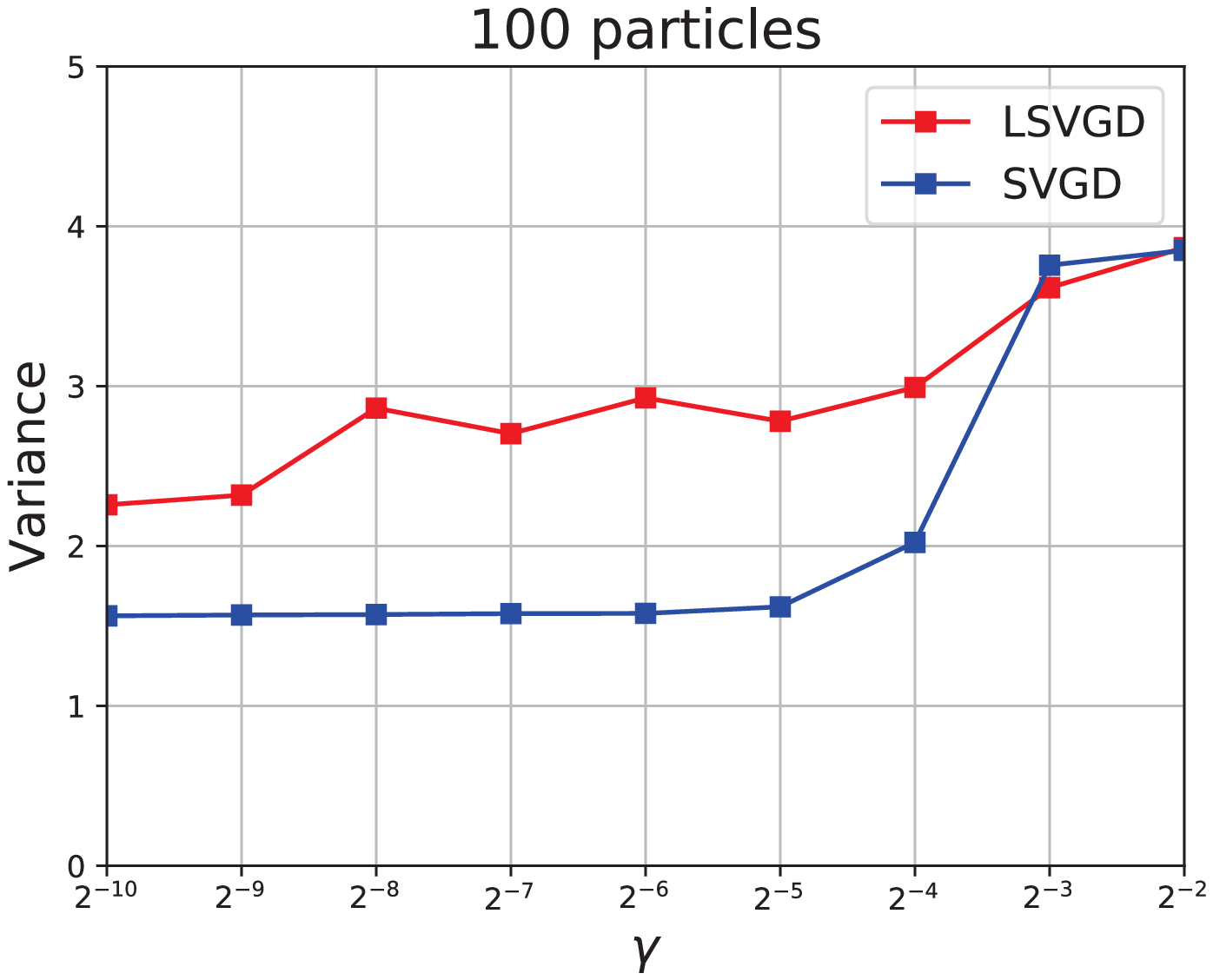}
\label{fig:var100}}
\caption{KL-divergence and variance of SVGD and LSVGD with different bandwidth $\gamma$ (from $2^{-10}$ to $2^{-2}$) and number of particles (20, 50, 100). }
\label{fig:toy2}
\end{figure*}
As an extension of Fig.~\ref{fig:toy}, we continue to investigate the behaviors of our LSVGD vs. standard SVGD on the same bimodal Gaussian distribution by varying the kernel parameter $\gamma$ from $2^{-10}$ to $2^{-2}$, and we test the performance under 20, 50 and 100 particles respectively. Each method takes 500 training iterations, and reports the final KL-divergence and variance.

Fig.~\ref{fig:toy2} shows the results that with the increase of $\gamma$ (i.e. the repulsive force goes stronger), we can observe that KL-divergence keeps decreasing for both LSVGD and SVGD. This is due to that larger $\gamma$ often yields larger spread of particles which leads to a better match between approximate and target distribution, and also raises the variance of particles of both methods. However, we can see that in all the three cases of different number of particles, LSVGD yields much lower KL-divergence under a wide range of parameter changing. This result of LSVGD outperforming SVGD is because it is less sensitive to kernel parameter, and it also empirically verifies LSVGD's convergence performance as analysed in Theorem 1. Additionally, LSVGD consistently yields higher variance than SVGD, which empirically verifies Theorem 2.


\subsection{Experiments on Cifar-10 Dataset}
\begin{table*}[t!]
\caption{Experimental results on Cifar-10 dataset.}
\begin{center}
\begin{tabular}{l|c|c|c|c|c|c|c|c|c}
\hline
\multirow{2}*{Algorithm} & \multicolumn{3}{|c}{Inception Score} & \multicolumn{3}{|c}{Accuracy (\%)} & \multicolumn{3}{|c}{Variance}\\
\cline{2-10}
~   & SGD &SVGD  &LSVGD & SGD &SVGD  &LSVGD & SGD &SVGD  &LSVGD\\
\hline
\hline
DCGAN \cite{radford2015unsupervised} (2015) &7.30$\pm$0.20 &7.22$\pm$0.22 &\textbf{7.52$\pm$0.25}      &75.31$\pm$1.05 &72.80$\pm$1.18 &\textbf{78.69$\pm$1.26}   &1.38 &1.39 &\textbf{1.44} \\
\hline
Stein-GAN \cite{wang2016learning} (2016)    &--   &7.35$\pm$0.22 &\textbf{7.77$\pm$0.24}      &--    &75.75$\pm$1.11 &\textbf{79.90$\pm$1.25}   &--   &1.46 &\textbf{1.63}\\
\hline
WGAN \cite{arjovsky2017wasserstein} (2017)  &7.55$\pm$0.18 &7.51$\pm$0.22 &\textbf{7.64$\pm$0.23}      &78.77$\pm$0.87 &78.63$\pm$1.02 &\textbf{80.83$\pm$1.07}   &1.42 &\textbf{1.49} &1.48 \\
\hline
WGAN-GP \cite{gulrajani2017improved} (2017) &7.88$\pm$0.20 &8.22$\pm$0.21 &\textbf{8.74$\pm$0.21}      &85.46$\pm$0.96 &89.36$\pm$1.14 &\textbf{93.30$\pm$1.20}   &1.46 &1.50 &\textbf{1.65} \\
\hline
Reg-GAN \cite{roth2017stabilizing} (2017)   &7.78$\pm$0.19 &7.85$\pm$0.24 &\textbf{8.92$\pm$0.27}      &83.21$\pm$1.02 &83.88$\pm$1.16 &\textbf{93.54$\pm$1.23}   &1.47 &1.43 &\textbf{2.00} \\
\hline
SNGAN \cite{miyato2018spectral} (2018)      &8.02$\pm$0.21 &8.08$\pm$0.23 &\textbf{8.75$\pm$0.25}      &86.65$\pm$1.03 &87.90$\pm$1.18 &\textbf{93.70$\pm$1.25}   &1.41 &1.43 &\textbf{1.69} \\
\hline
SAGAN \cite{zhang2019self} (2019)           &7.80$\pm$0.20 &8.19$\pm$0.24 &\textbf{8.87$\pm$0.26}      &83.69$\pm$1.05 &86.39$\pm$1.20 &\textbf{92.32$\pm$1.28}   &1.43 &1.65 &\textbf{2.16}  \\
\hline
\end{tabular}
\end{center}
\label{tb:cifarres}
\end{table*}

\begin{figure*}[t!]
\center
\subfigure[]{\includegraphics[width=0.26\textwidth]{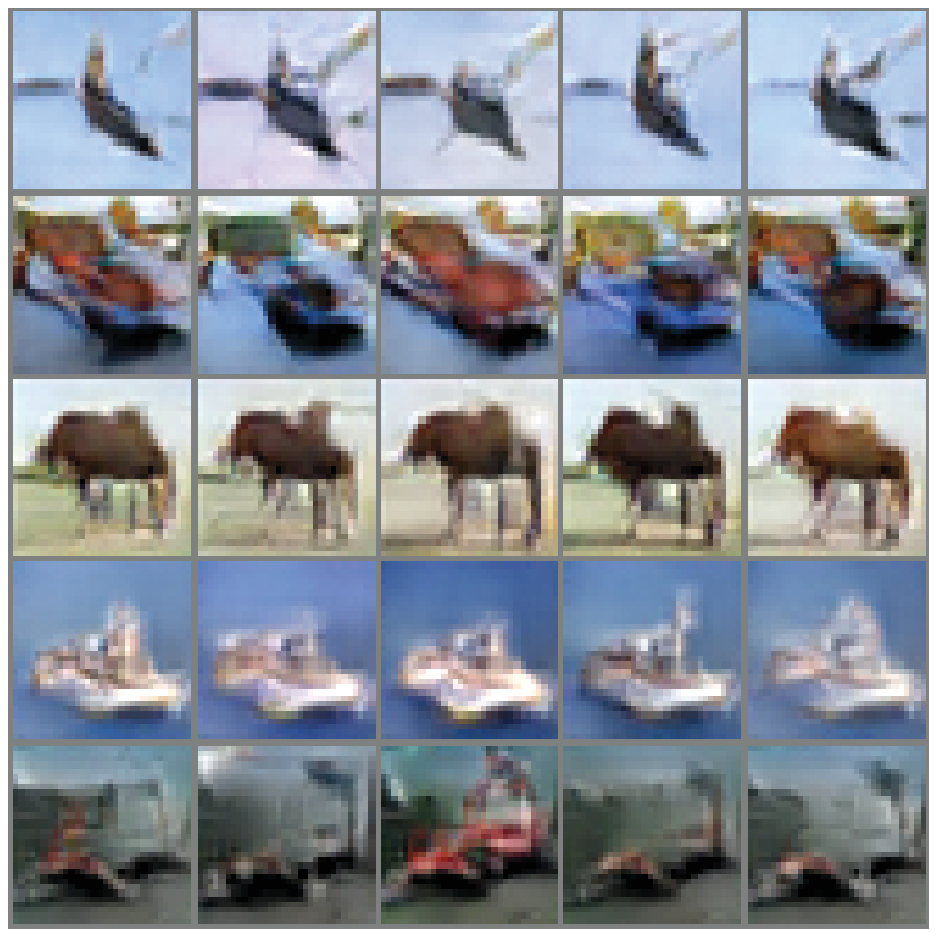}
\label{fig:sgdimg}}
\subfigure[]{\includegraphics[width=0.26\textwidth]{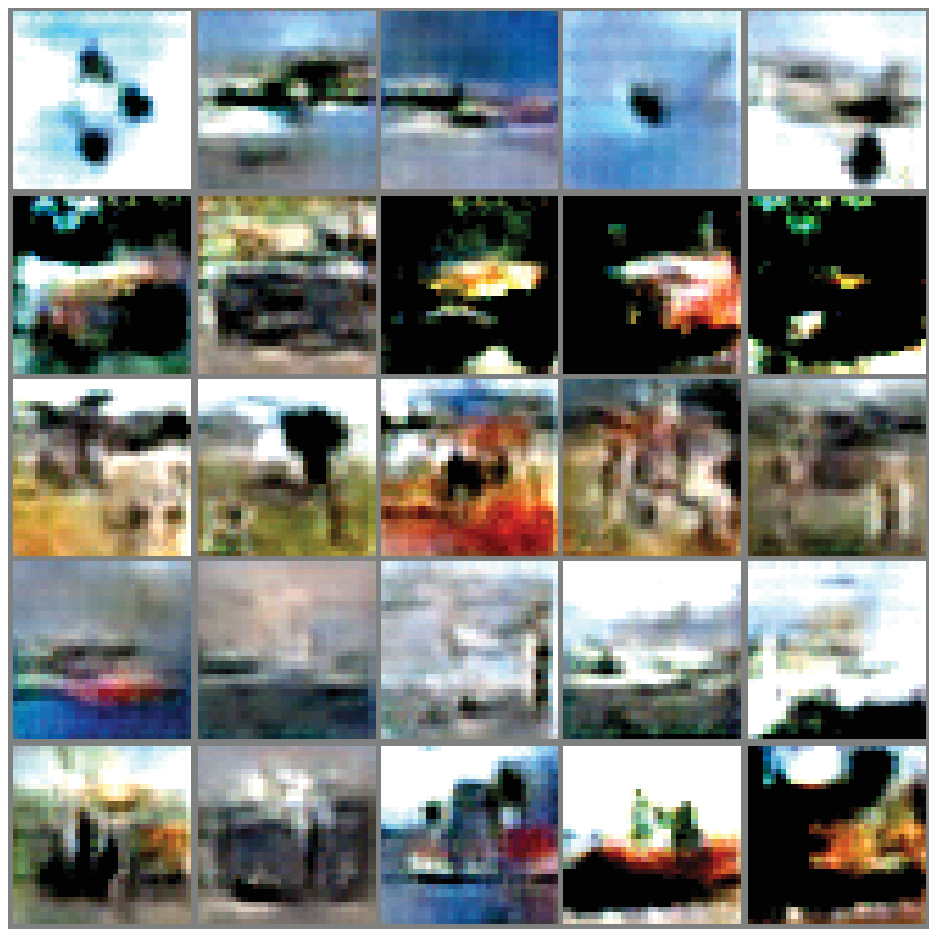}
\label{fig:svgdimg}}
\subfigure[]{\includegraphics[width=0.26\textwidth]{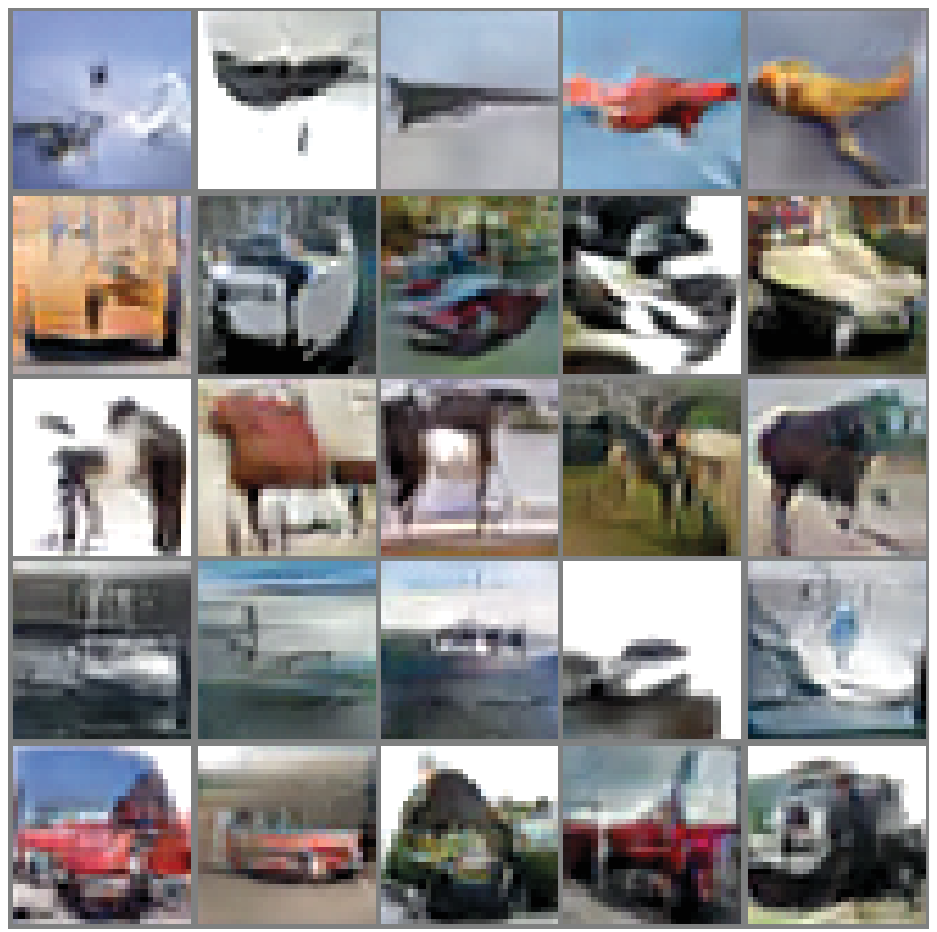}
\label{fig:lsvgdimg}}
\subfigure[]{\includegraphics[width=0.27\textwidth]{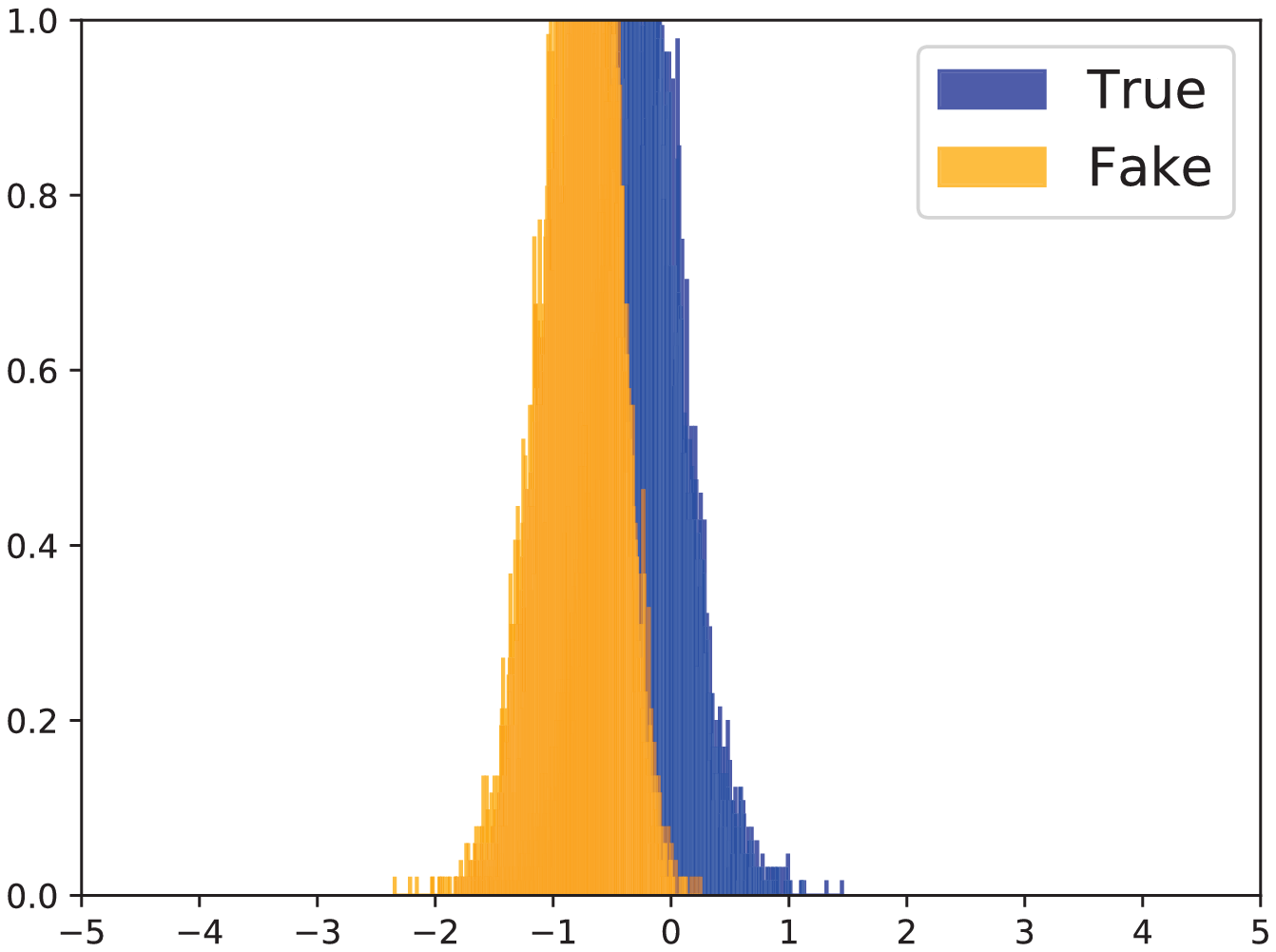}
\label{fig:sgdvar}}
\subfigure[]{\includegraphics[width=0.27\textwidth]{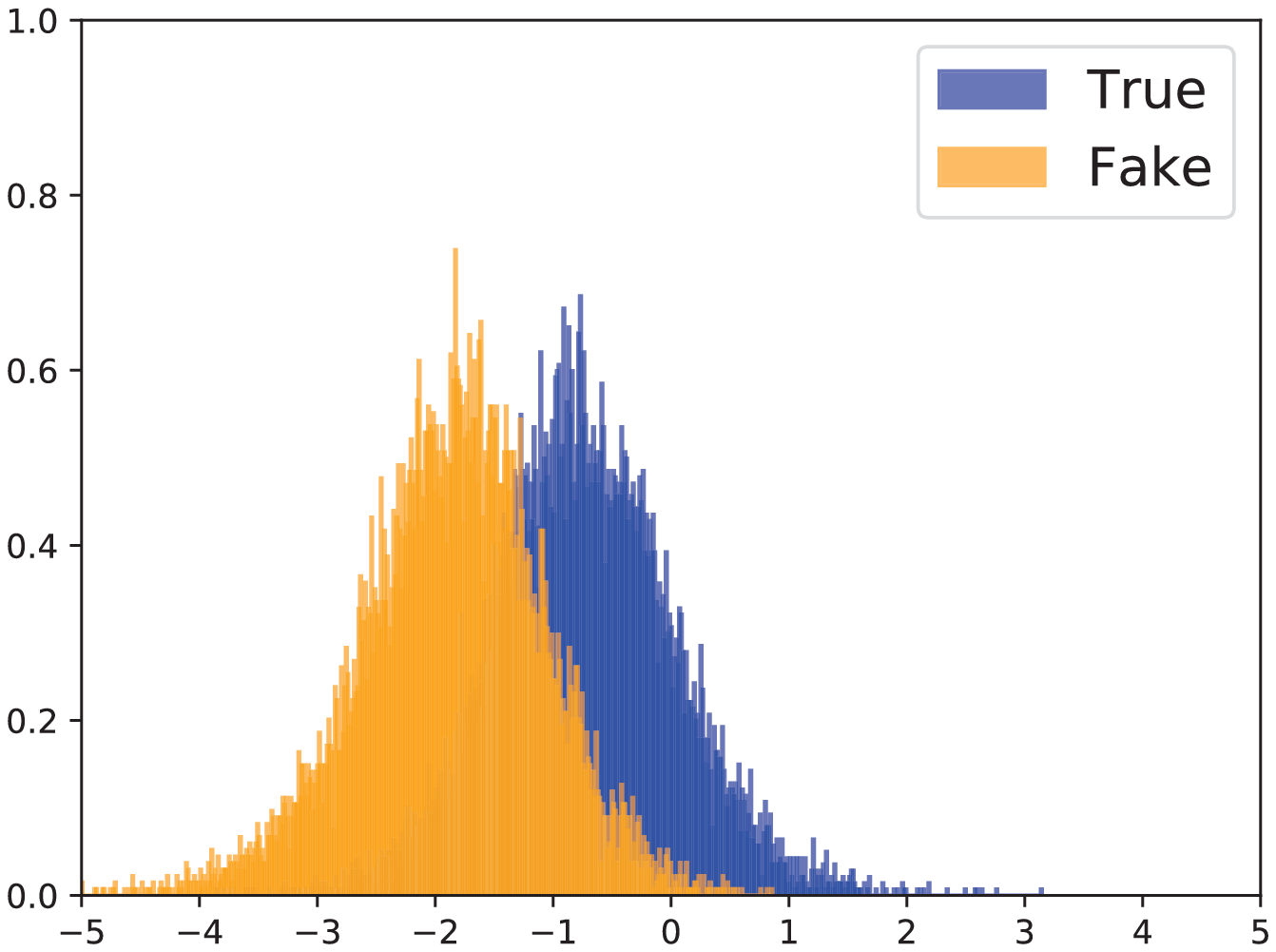}
\label{fig:svgdvar}}
\subfigure[]{\includegraphics[width=0.27\textwidth]{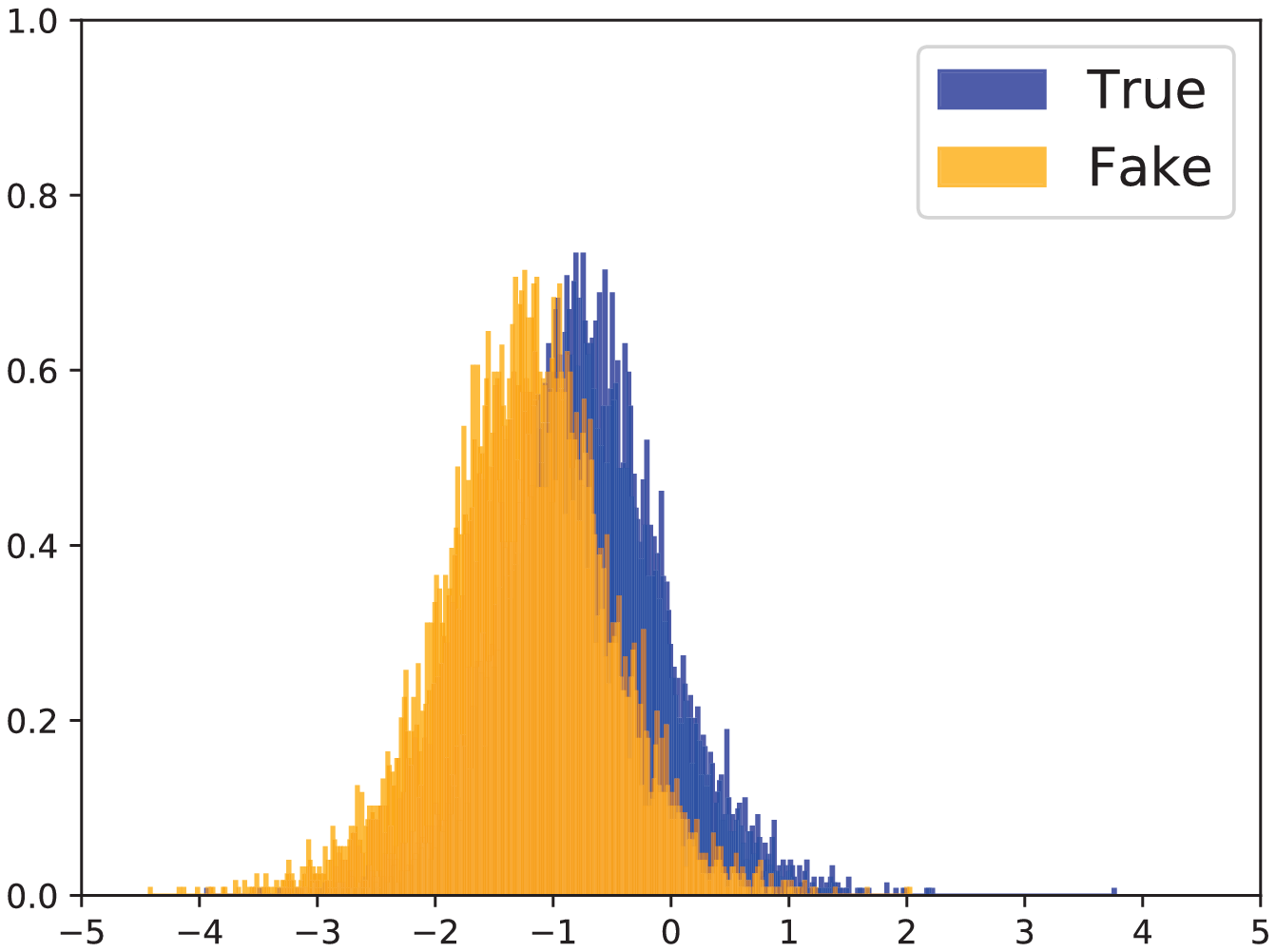}
\label{fig:lsvgdvar}}
\caption{Experimental results of DCGAN. The first row shows the generated images of training with (a) SGD, (b) SVGD, (c) LSVGD. The second row shows the empirical distribution of 10K real test images (true) and 10K generated images of (d) SGD, (e) SVGD, (f) LSVGD respectively.}
\label{fig:cifar}
\end{figure*}

\begin{figure*}[t!]
\center
\subfigure[]{\includegraphics[width=0.28\textwidth]{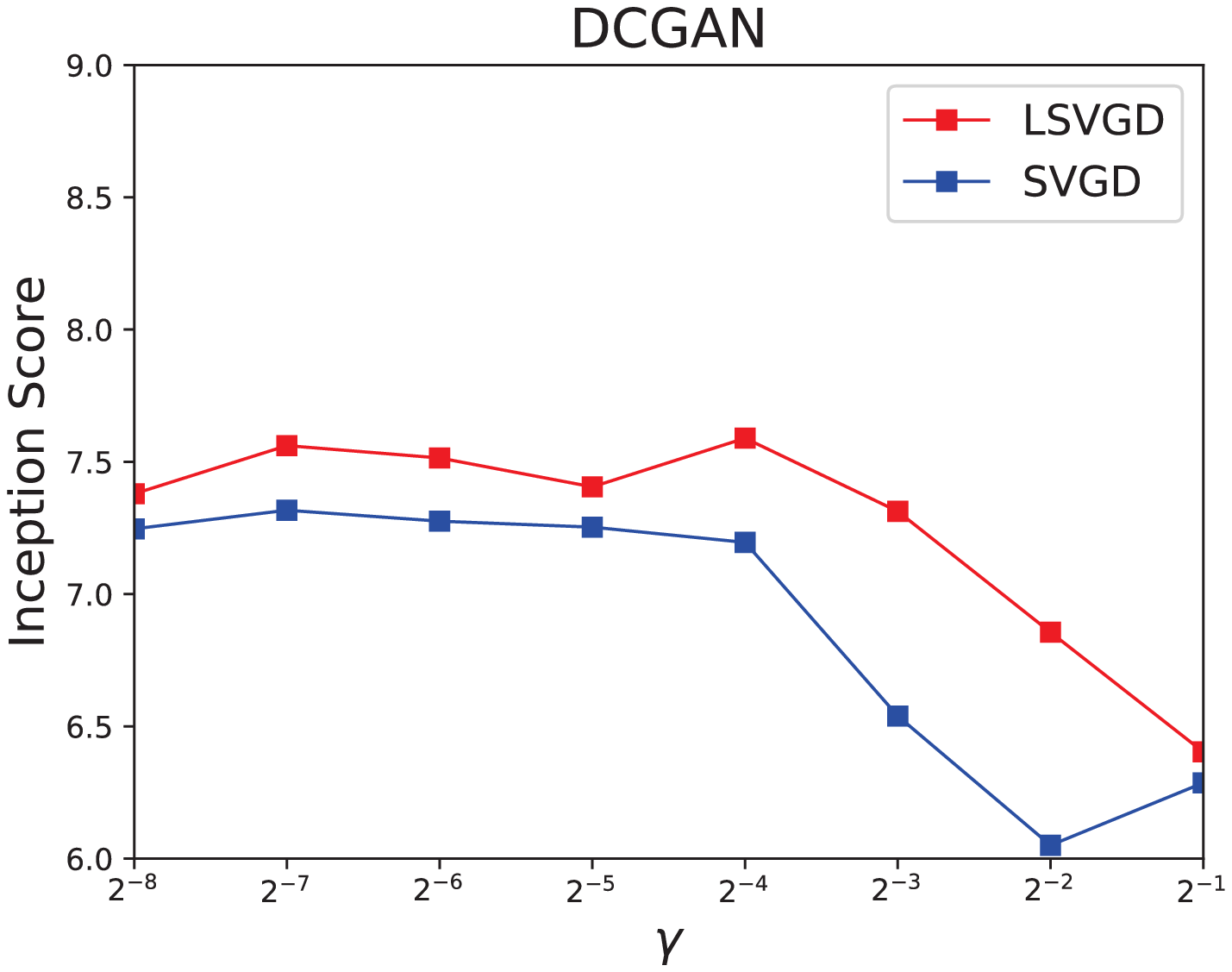}
\label{fig:dckis}}
\subfigure[]{\includegraphics[width=0.28\textwidth]{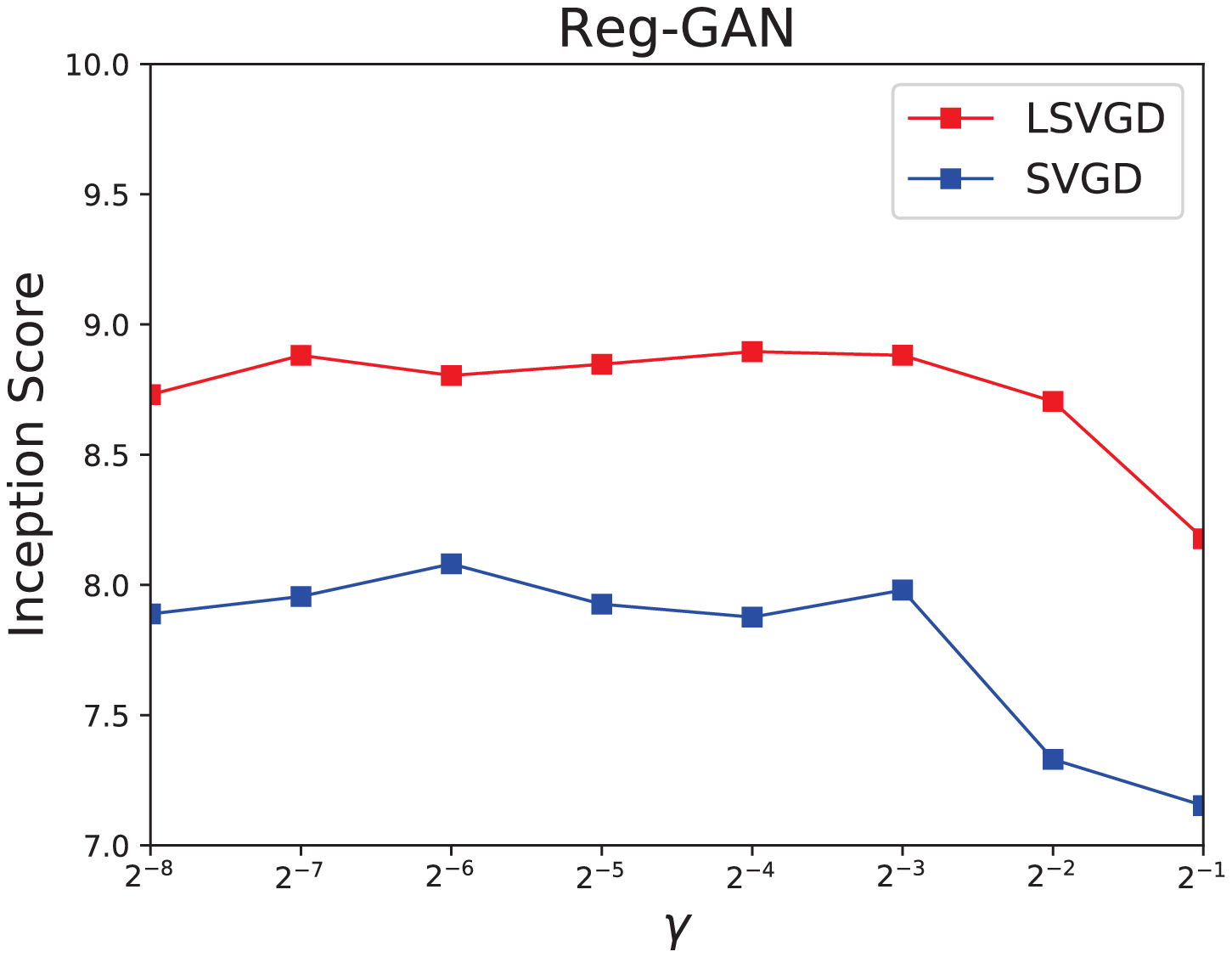}
\label{fig:gpkis}}
\subfigure[]{\includegraphics[width=0.28\textwidth]{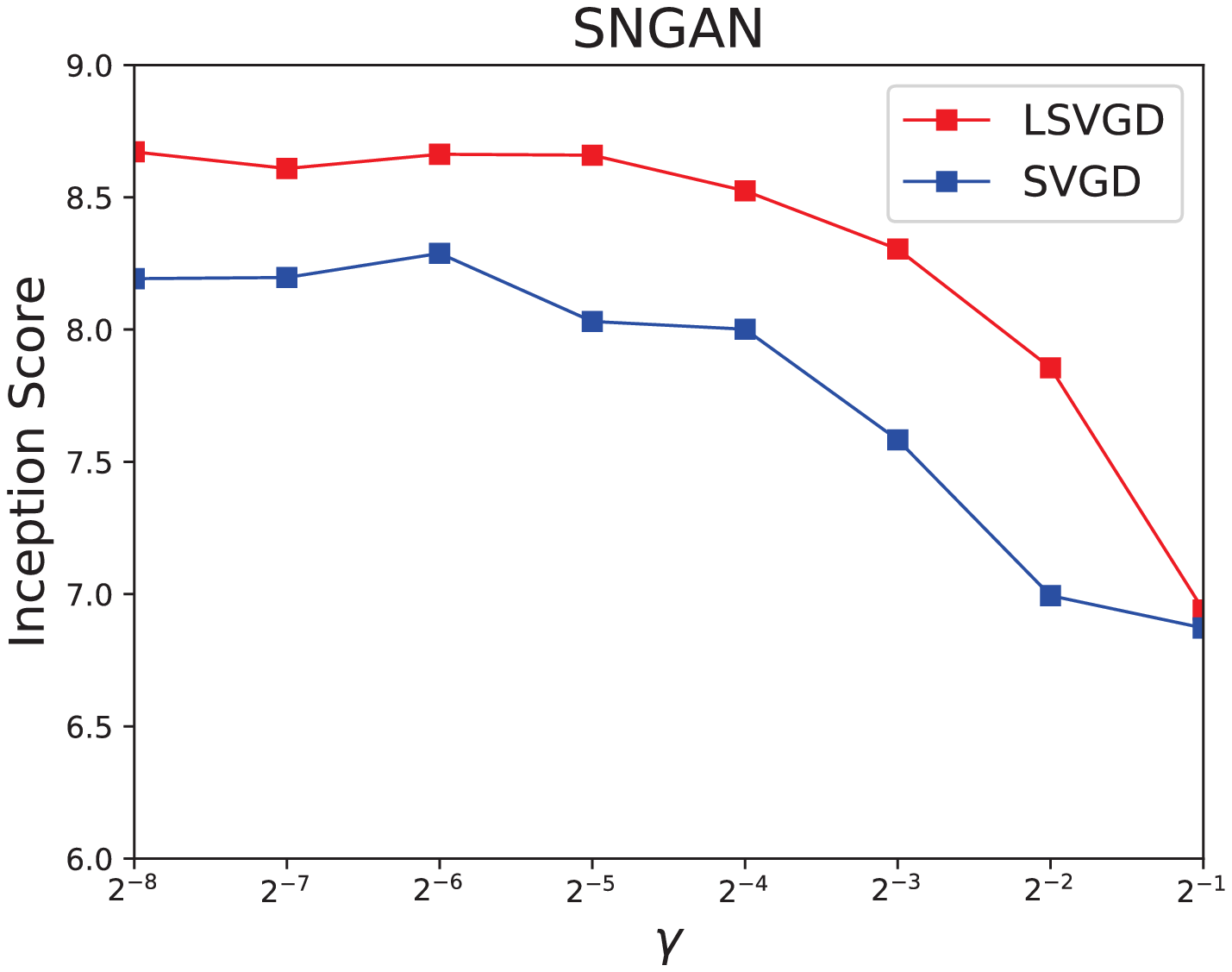}
\label{fig:snkis}}
\subfigure[]{\includegraphics[width=0.28\textwidth]{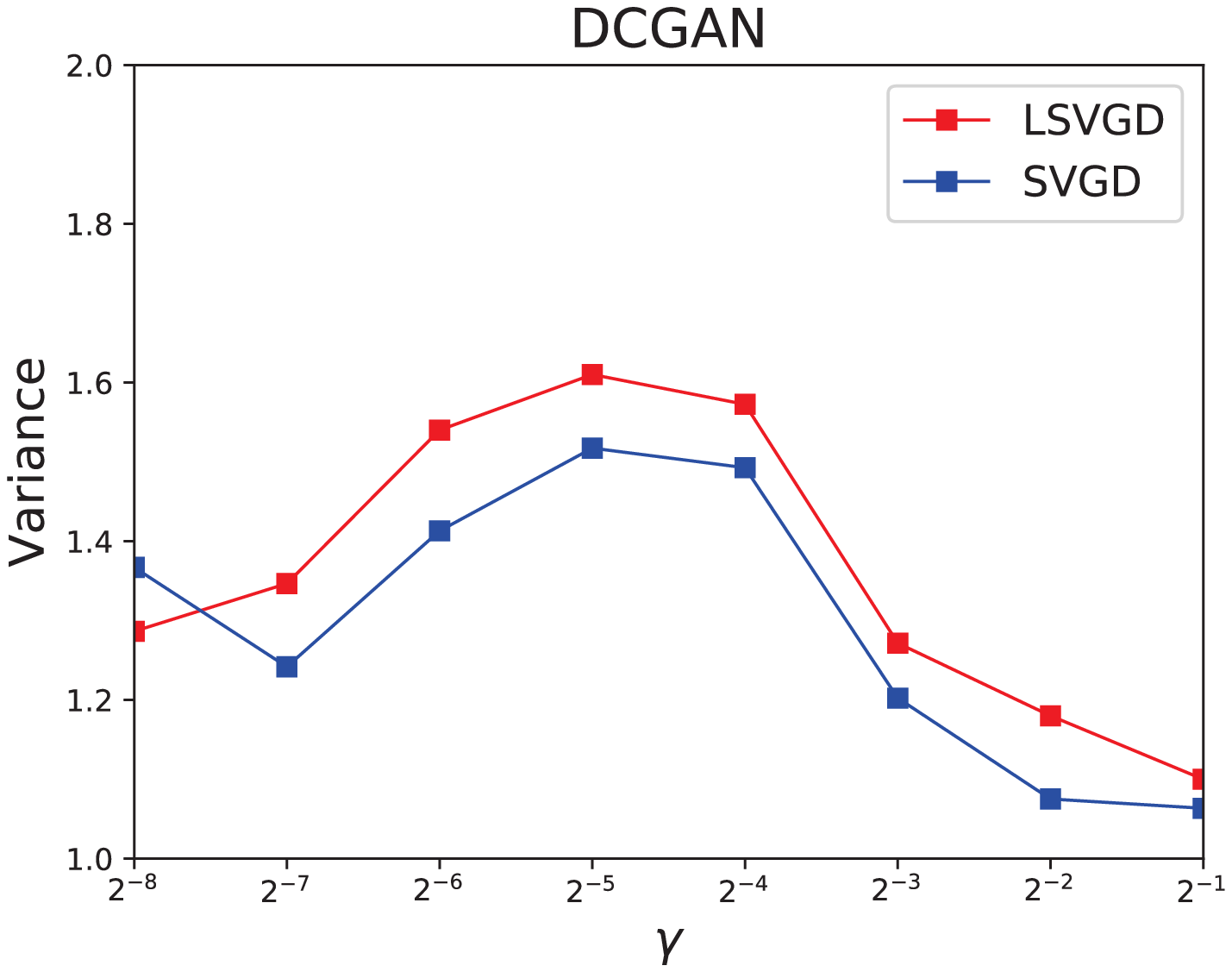}
\label{fig:dckvar}}
\subfigure[]{\includegraphics[width=0.28\textwidth]{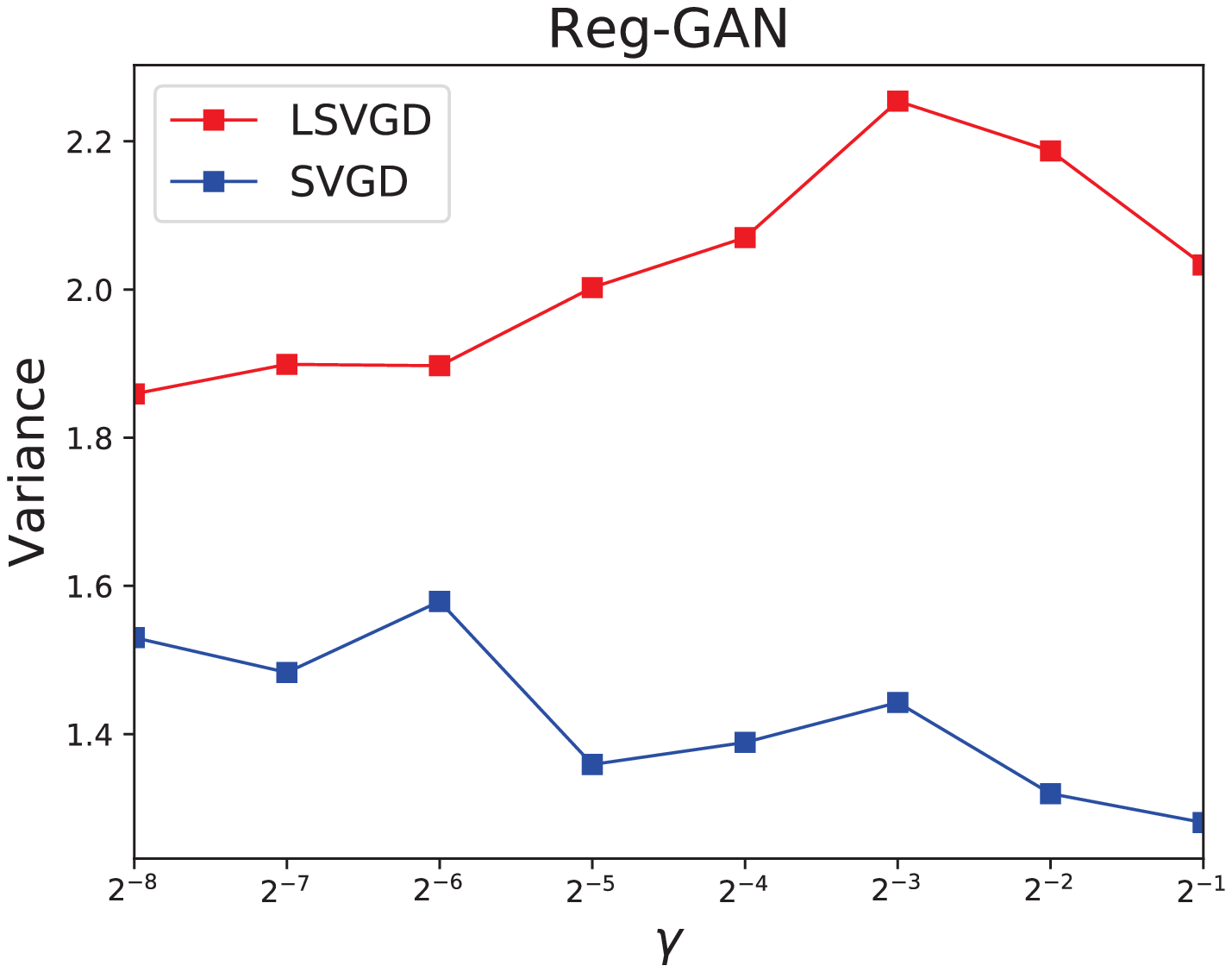}
\label{fig:gpkvar}}
\subfigure[]{\includegraphics[width=0.28\textwidth]{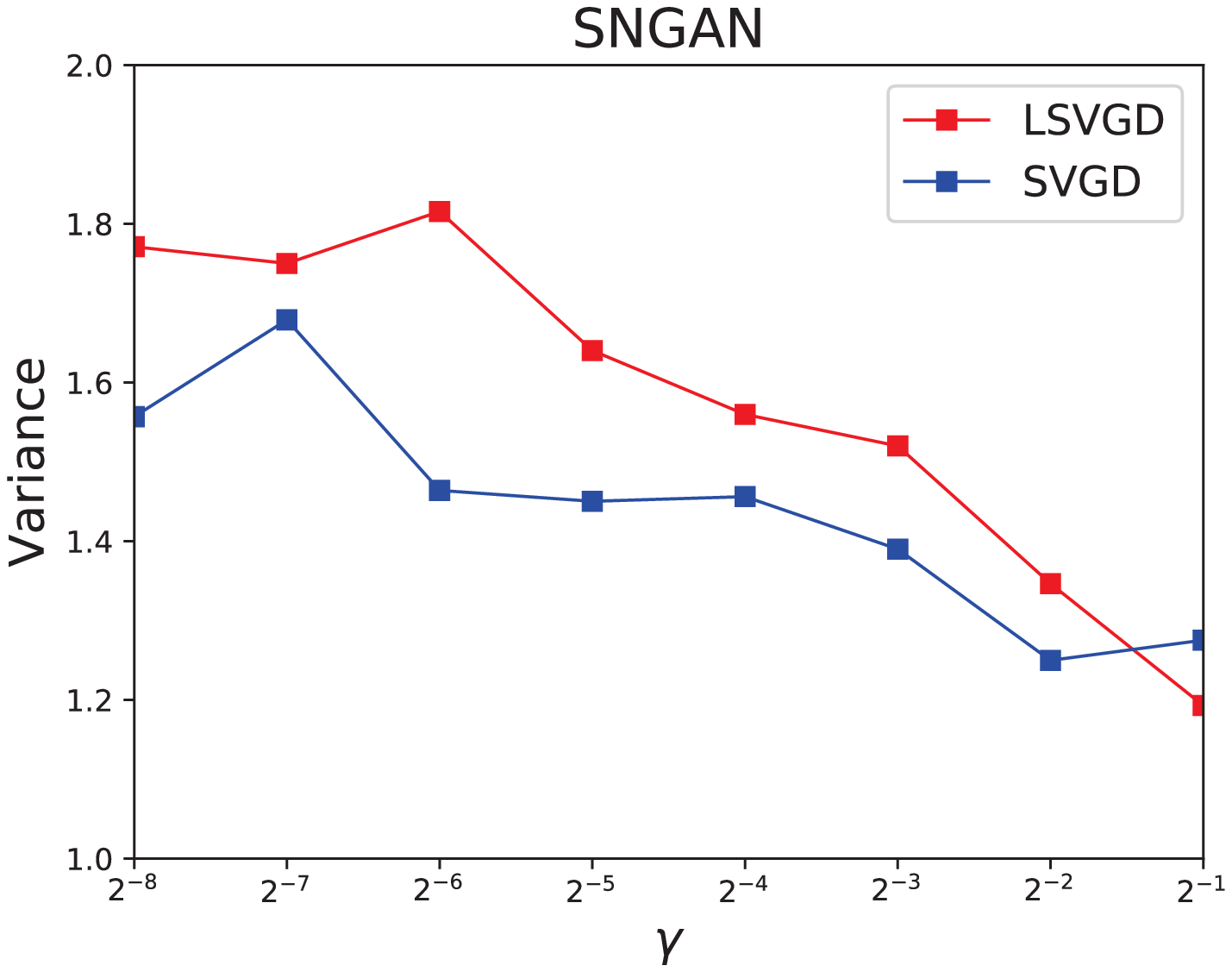}
\label{fig:snkvar}}
\caption{The effect of kernel parameter. (a), (b), (c) are respectively the results of inception score of DCGAN, Reg-GAN and SNGAN, (d), (e), (f) are respectively the results of variance of DCGAN, Reg-GAN and SNGAN. }
\label{fig:expkl}
\end{figure*}

\begin{figure*}[t!]
\center
\subfigure[]{\includegraphics[width=0.28\textwidth]{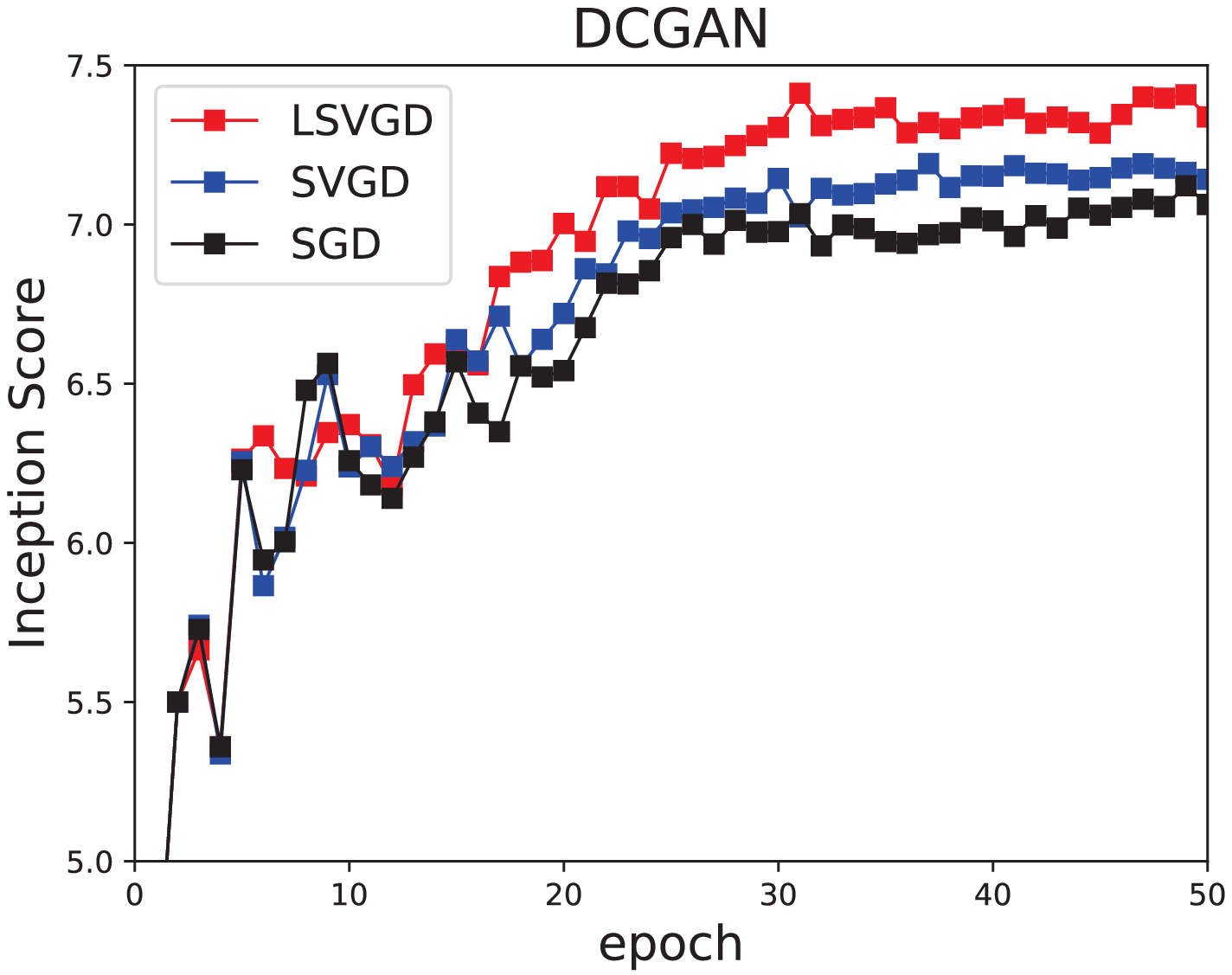}
\label{fig:dcgan}}
\subfigure[]{\includegraphics[width=0.28\textwidth]{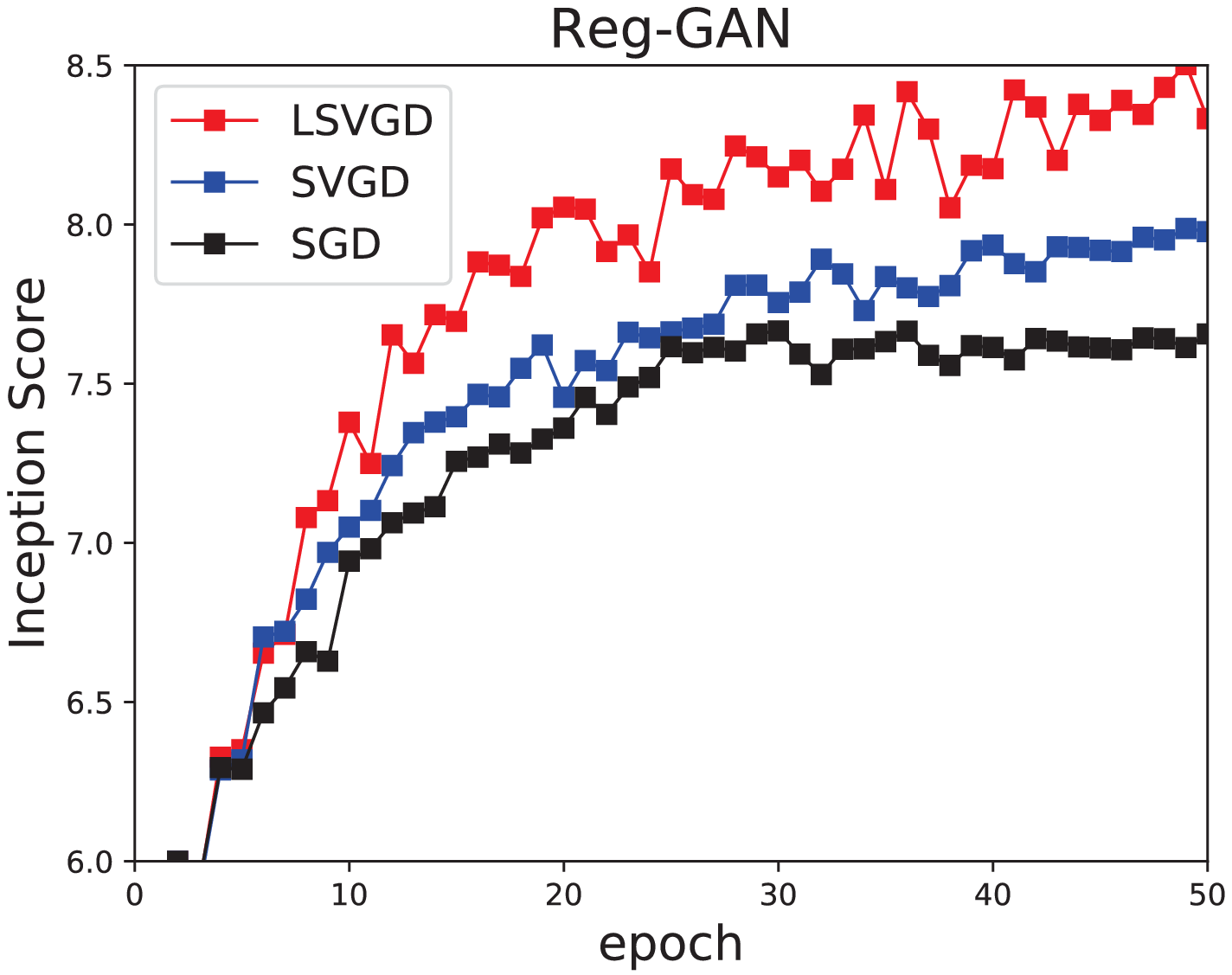}
\label{fig:reggan}}
\subfigure[]{\includegraphics[width=0.28\textwidth]{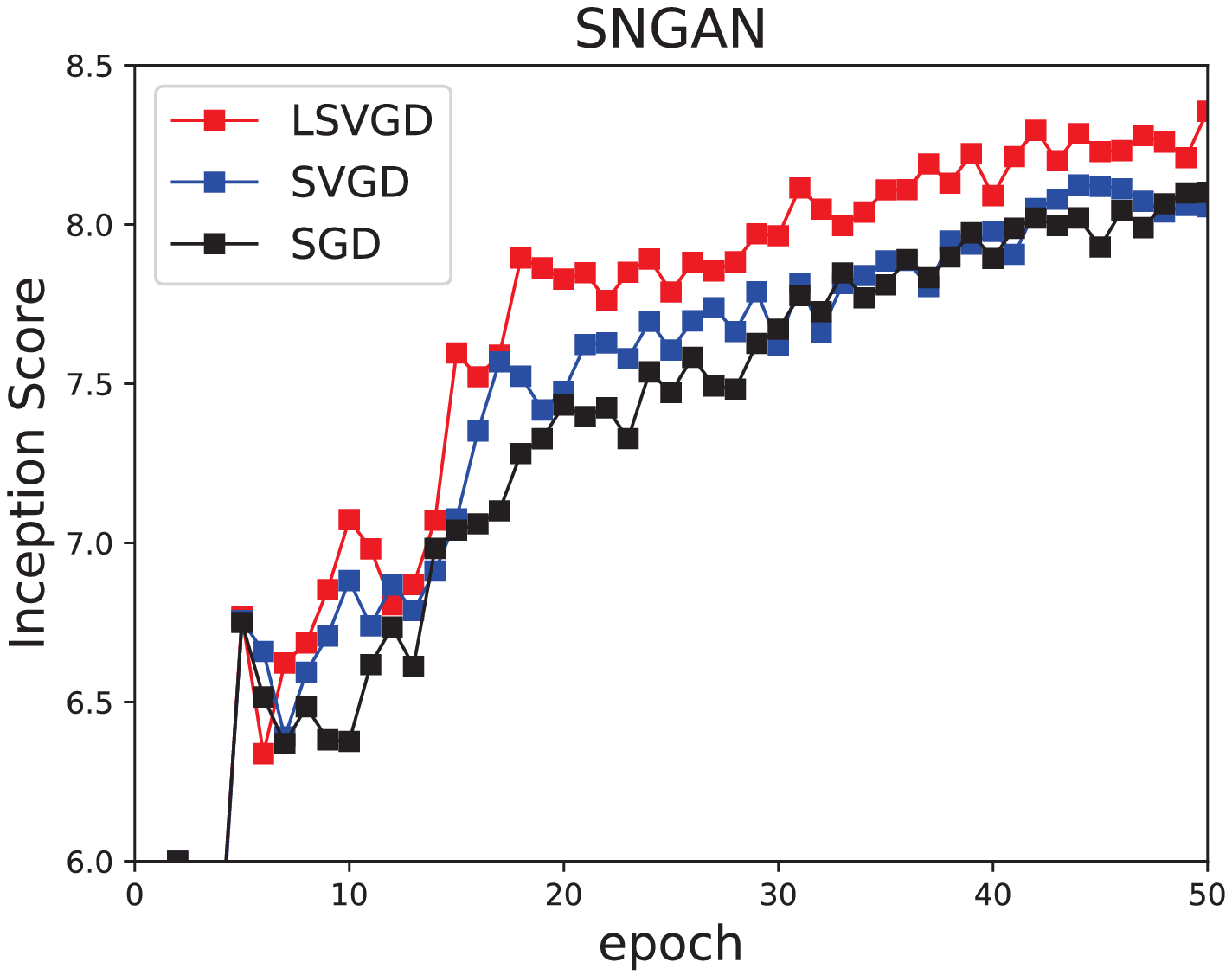}
\label{fig:sngan}}
\caption{How inception score varies as training epoch goes. (a), (b), (c) are respectively the results of DCGAN, Reg-GAN and SNGAN trained with three methods. }
\label{fig:expiter}
\end{figure*}

Cifar-10 \cite{krizhevsky2009learning} is a popular benchmark dataset containing 60k $32\times 32$ RGB images of 10 categories: airplane, automobile, bird, cat, deer, dog, frog, horse, ship, truck. A standard protocol on this dataset divides the entire dataset into a 50k training set and 10k test set. We conduct a serial of experiments on this dataset by training various GAN models with original stochastic gradient descent (denoted as SGD), SVGD and LSVGD.

\subsubsection{Parameter setting}
All the GAN models follow the basic net structure as \cite{radford2015unsupervised} except that 1) the input layer of generator is a concatenation of a random noise vector and a random label vector. 2) For the sake of efficiency, we insert a compact feature layer with dimensionality of 24 before the outermost layer of discriminator, where we replace original gradient with LSVGD or SVGD. The batch size (i.e., the number of particles) is set to 100. All the methods go through 50 training epochs with a fixed learning rate of $10^{-4}$. For selecting the fittest kernel parameter, we randomly sample 5000 images from the training set as a validation set for running grid-search.

\subsubsection{Performance Evaluation}
On this dataset, We apply LSVGD on 7 GAN models including 2 baseline models: DCGAN \cite{radford2015unsupervised} and Stein-GAN \cite{wang2016learning} \footnote{The Stein-GAN architecture directly follows \cite{wang2016learning}.}, and 5 regularized models: WGAN \cite{arjovsky2017wasserstein}, WGAN-GP \cite{gulrajani2017improved}, Reg-GAN \cite{roth2017stabilizing}, SNGAN \cite{miyato2018spectral} and SAGAN \cite{zhang2019self}.

We test the performance of GAN models under three evaluation metrics: inception score \cite{barratt2018note}, classification accuracy and variance. Inception score is the most commonly used metric in prior works. However as pointed out in \cite{barratt2018note}, a finetuned model (on the current dataset) should be used instead for more precise evaluation. Therefore, we work with a finetuned residual network \cite{he2016deep}, i.e., pretrained on ImageNet dataset and finetuned on Cifar-10. We let each GAN model generate 10K images and calculate the inception score based on the output of the evaluation model. In addition, we also report the classification accuracy and intra-class variance of the generated images. The variance is also calculated with the features extracted by the evaluation network.

For each of those compared models, we report the performance under 3 settings: training with SGD, SVGD and LSVGD. We repeat each method for 5 times and report the average and standard deviation. Experimental results are shown in Table.~\ref{tb:cifarres}, from which we have three basic observations:

First, in terms of inception score, training with SVGD does not always promise a performance gain compared with original gradient descent. One can see that SVGD achieves 7.22 on DCGAN which is lower than the 7.30 of original gradient descent, and 7.51 on WGAN lower than the original 7.55. Similarly, for classification accuracy, SVGD achieves 72.80\% on DCGAN lower than the original 75.31\%, and 78.63\% on WGAN lower than the original 78.77\%. SVGD performs better than original gradient descent on the other GAN models (except for Stein-GAN, since this model is originally designed for SVGD based training), however the performance gain on Reg-GAN and SNGAN is quite marginal. This results exhibit the instability of SVGD.

Second, on all the methods compared, LSVGD consistently outperforms SGD and SVGD with a statistical significance ($p$ = 0.05) in terms of both inception score and classification accuracy. Notably, even for those regularized GANs such as WGAN-GP, Reg-GAN, SNGAN and SAGAN, LSVGD still performs better than SVGD under both of the metrics. This verifies LSVGD's ability of enhancing particle momentum which allows particles to reach better local optimums.

Last, we can see from the last column of Table.~\ref{tb:cifarres} LSVGD almost achieves larger variance of particles than original gradient descend and SVGD except on WGAN where SVGD achieves slightly higher variance than LSVGD. This is probably due to the inherent regularizer of weight clipping imposed on WGAN. Surprisingly, LSVGD achieves a variance of 2.00 on Reg-GAN and 2.16 on SAGAN which significantly improves SVGD whose variance is 1.43 and 1.65 on Reg-GAN and SAGAN respectively. These results show that LSVGD not only yields higher diversity than SVGD but also improves the image quality which is spoken out by both inception score and classification accuracy.

\subsubsection{Addressing mode collapse issue}
We take for an instance DCGAN to illustrate how LSVGD handles the mode collapse issue. We train three models respectively with SGD, SVGD and LSVGD. The kernel parameter $\gamma$ is set to 0.05 for both SVGD and LSVGD respectively. Fig.~\ref{fig:sgdimg} shows the generated images of original training. We can see that mode collapse happens when training with SGD due to the lack of repulsive force among particles. This can be further verified by comparing the empirical distribution of 10K real test images and 10K generated images.\footnote{Note that for better illustration, this result is based on the 1-dimensional output value of discriminator (i.e., the real-fake classification) instead of on images space.} We can see from Fig.~\ref{fig:sgdvar} that the resulting images have fairly small variance which indicates the low diversity of generated images. Compared with SGD, SVGD can enhance diversity (Fig.~\ref{fig:svgdvar}) with its particle-based mechanism. However, its over-sensitivity to parameter probably results in bad match between synthesized and real images, which is also a cause of blurred images as illustrated in Fig.~\ref{fig:svgdimg}. Fortunately, the results of LSVGD (Fig.~\ref{fig:lsvgdimg} and Fig.~\ref{fig:lsvgdvar}) show that it not only enhances the diversity of images, but generates sharp and realistic-looking images.

\subsubsection{The effect of kernel parameter}

Moreover, we test the parameter sensitivity of LSVGD and SVGD on three GAN models, one baseline model: DCGAN and two regularized models: Reg-GAN and SNGAN. We vary $\gamma$ between $2^{-8}$ and $2^{-1}$. Fig.~\ref{fig:dckis} to Fig.~\ref{fig:snkis} shows the result of inception score. One can see that SVGD and LSVGD have similar performance on all the three GAN models, which keeps stable within the range $\gamma \le 2^{-4}$ and then decreases after $2^{-4}$. In terms of inception score, LSVGD performs consistently better over nearly the entire range of $\gamma$. Fig.~\ref{fig:dckvar} to Fig.~\ref{fig:snkvar} shows the result of variance that indicates the diversity of generated images. First, one can see that variance is more sensitive to $\gamma$, but LSVGD effectively enhances variance on all the three GAN models. Second, for both DCGAN and SNGAN, large $\gamma$ can also lead to low variance which is different from the results on toy data (Fig.~\ref{fig:toy2}). This is probably because improper $\gamma$ would lead to bad synthesis which is not able to cheat discriminator or acquire valid feedback. Hence, if the generators is not properly trained, then it can not generate diversified images. Additionally, Reg-GAN is more robust to parameter selection due to its inherent regularization. However even in this case, LSVGD can still largely enhance the diversity (variance) over the entire range of $\gamma$. This experimental results demonstrate effectiveness of LSVGD as an useful enhancement for GAN training.

\subsubsection{Training curves}

We also verify the convergence property of our method by investigating how inception score changes as training epoch goes. We train all the three models from scratch . Fig.~\ref{fig:expiter} shows the experimental result that LSVGD also has good convergence property and can perform better than SGD and SVGD, even though LSVGD's result fluctuates a little more over iterations which can be attributed to the injected noise. On the other hand, adding extra disturbance allows particles to explore for better local optimums.

\begin{table*}[t!]
\caption{Experimental results on Tiny-ImageNet dataset.}
\begin{center}
\begin{tabular}{l|c|c|c|c|c|c|c|c|c}
\hline
\multirow{2}*{Algorithm} & \multicolumn{3}{|c}{Inception Score} & \multicolumn{3}{|c}{Accuracy (\%)} & \multicolumn{3}{|c}{Variance}\\
\cline{2-10}
~   & SGD &SVGD  &LSVGD & SGD &SVGD  &LSVGD & SGD &SVGD  &LSVGD\\
\hline
\hline
DCGAN \cite{radford2015unsupervised} (2015) &31.51$\pm$0.21 &32.63$\pm$0.25 &\textbf{35.76$\pm$0.28}      &21.56$\pm$1.07 &20.32$\pm$1.20 &\textbf{27.00$\pm$1.22}    &8.08 &8.65 &\textbf{8.81} \\
\hline
Stein-GAN \cite{wang2016learning} (2016)    &--    &33.57$\pm$0.26 &\textbf{36.13$\pm$0.28}      &--    &24.41$\pm$1.21 &\textbf{27.83$\pm$1.24}    &--   &8.72 &\textbf{8.92}\\
\hline
WGAN \cite{arjovsky2017wasserstein} (2017)  &36.68$\pm$0.19 &36.95$\pm$0.23 &\textbf{40.06$\pm$0.27}      &37.35$\pm$1.02 &35.97$\pm$1.16 &\textbf{41.33$\pm$1.20}    &8.61 &8.46 &\textbf{8.70}\\
\hline
WGAN-GP \cite{gulrajani2017improved} (2017) &38.71$\pm$0.20 &49.54$\pm$0.24 &\textbf{50.79$\pm$0.27}      &36.81$\pm$1.05 &53.07$\pm$1.20 &\textbf{53.14$\pm$1.24}    &9.24 &9.23 &\textbf{9.43}\\
\hline
Reg-GAN \cite{roth2017stabilizing} (2017)   &39.88$\pm$0.19 &48.14$\pm$0.22 &\textbf{55.00$\pm$0.26}      &43.68$\pm$1.04 &53.28$\pm$1.19 &\textbf{58.57$\pm$1.24}    &8.98 &9.00 &\textbf{9.92} \\
\hline
SNGAN \cite{miyato2018spectral} (2018)      &57.74$\pm$0.22 &86.30$\pm$0.24 &\textbf{89.65$\pm$0.28}      &59.26$\pm$1.06 &76.77$\pm$1.21 &\textbf{78.37$\pm$1.26}    &10.11 &11.32 &\textbf{11.64} \\
\hline
SAGAN \cite{zhang2019self} (2019)           &41.03$\pm$0.23 &55.54$\pm$0.26 &\textbf{60.81$\pm$0.29}      &38.45$\pm$1.08 &58.28$\pm$1.24 &\textbf{63.56$\pm$1.28}    &9.46 &10.02 &\textbf{10.51}\\
\hline
\end{tabular}
\end{center}
\label{tb:tinyImg}
\end{table*}
\begin{table*}[t!]
\caption{Experimental results on CelebA dataset.}
\begin{center}
\begin{tabular}{l|c|c|c|c|c|c}
\hline
\multirow{2}*{Algorithm} & \multicolumn{3}{|c}{Accuracy (\%)} & \multicolumn{3}{|c}{Variance} \\
\cline{2-7}
~   & SGD &SVGD  &LSVGD & SGD &SVGD  &LSVGD \\
\hline
\hline
DCGAN \cite{radford2015unsupervised} (2015) &\textbf{71.37$\pm$0.96} &66.76$\pm$1.08 &71.24$\pm$1.12      &2.24 &2.26 &\textbf{2.27} \\
\hline
Stein-GAN \cite{wang2016learning} (2016)    &--   &69.98$\pm$1.10 &\textbf{72.55$\pm$1.13}      &--    &2.14 &\textbf{2.19} \\
\hline
WGAN \cite{arjovsky2017wasserstein} (2017)  &69.51$\pm$0.88 &68.74$\pm$1.01 &\textbf{69.60$\pm$1.03}      &2.18 &2.17 &\textbf{2.23} \\
\hline
WGAN-GP \cite{gulrajani2017improved} (2017) &\textbf{76.17$\pm$0.97} &75.43$\pm$1.07 &75.68$\pm$1.11      &2.11 &2.16 &\textbf{2.25} \\
\hline
Reg-GAN \cite{roth2017stabilizing} (2017)   &79.66$\pm$0.99 &78.36$\pm$1.07 &\textbf{80.26$\pm$1.10}      &1.96 &1.90 &\textbf{2.04} \\
\hline
SNGAN \cite{miyato2018spectral} (2018)      &78.79$\pm$1.01 &79.24$\pm$1.09 &\textbf{79.90$\pm$1.12}      &2.02 &2.12 &\textbf{2.24} \\
\hline
SAGAN \cite{zhang2019self} (2019)           &81.45$\pm$1.03 &80.72$\pm$1.10 &\textbf{81.91$\pm$1.14}      &2.06 &2.04 &\textbf{2.12} \\
\hline
\end{tabular}
\end{center}
\label{tb:CelebAKL}
\end{table*}

\begin{figure*}[t!]
\center
\subfigure[]{\includegraphics[width=0.28\textwidth]{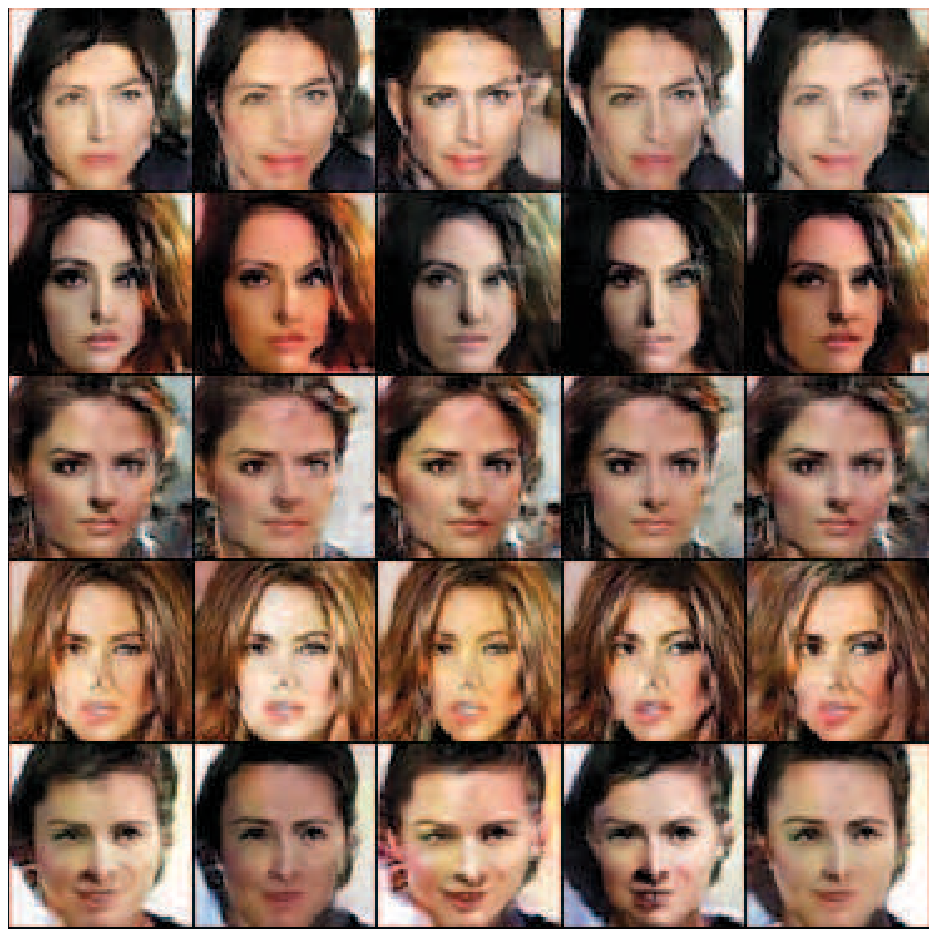}
\label{fig:celebm}}
\subfigure[]{\includegraphics[width=0.28\textwidth]{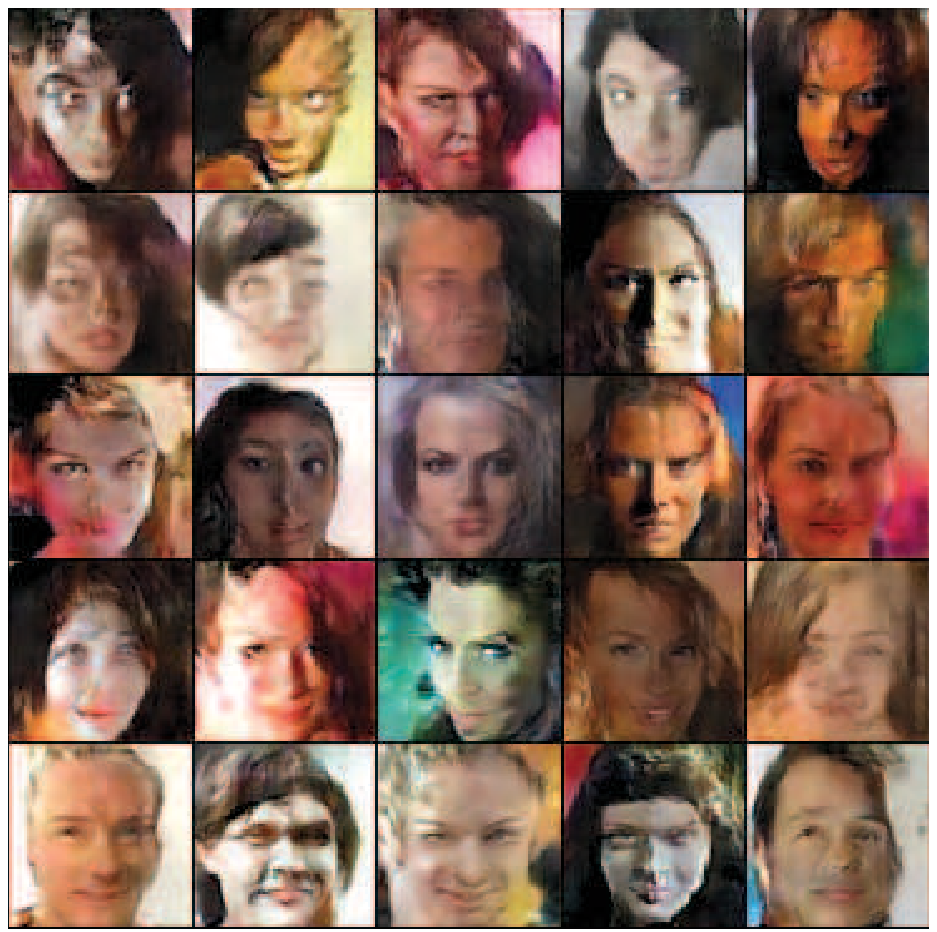}
\label{fig:celebs}}
\subfigure[]{\includegraphics[width=0.28\textwidth]{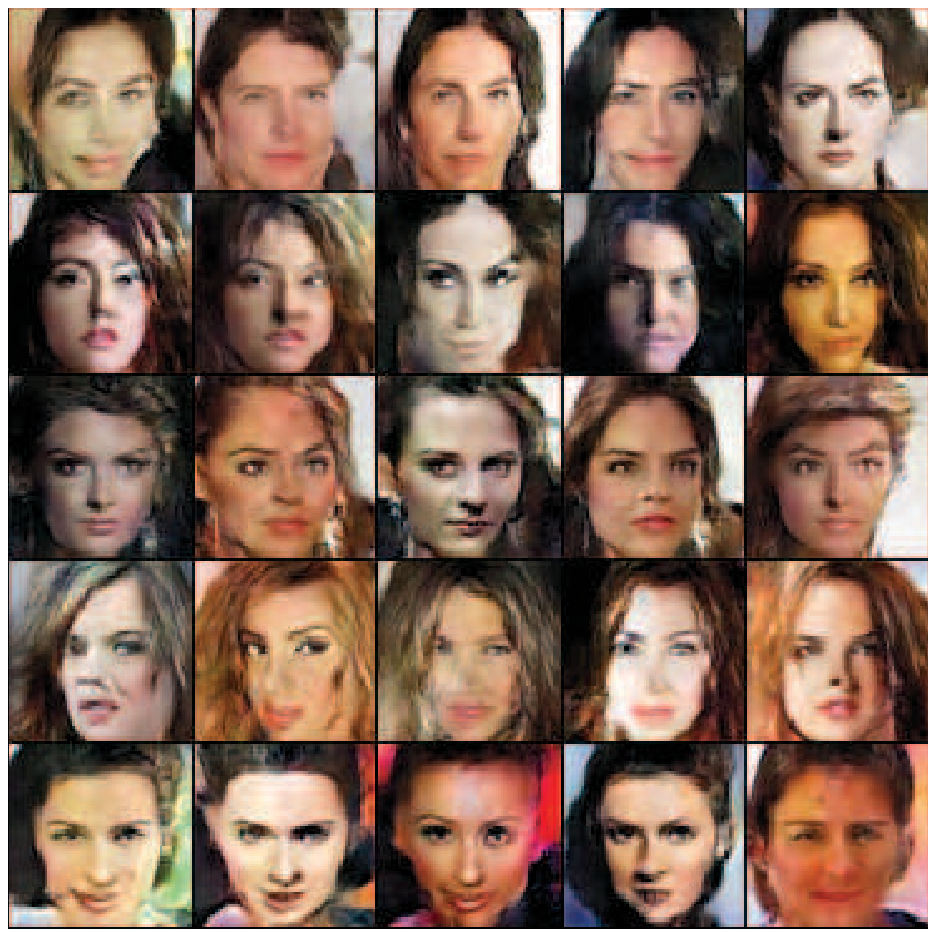}
\label{fig:celebl}}
\caption{Generated images of CelebA dataset with (a) SGD, (b) SVGD, (c) LSVGD respectively.}
\label{fig:celebImg}
\end{figure*}

\subsection{Experiments on Tiny-ImageNet Dataset}
Tiny-Imagenet \cite{russakovsky2015imagenet} is subset of the large scale dataset ImageNet \cite{deng2009imagenet}, which has 200 classes with each class containing 500 training images and 50 validation images. All the images are cropped and resized to 64 $\times$ 64. All the GAN models follow the same structure as \cite{radford2015unsupervised} with generated image size being 64 $\times$ 64. The Stein-GAN follows its own structure of \cite{wang2016learning}. We still use a finetuned ResNet model for evaluation. The other settings are the same as Cifar-10.

Table.~\ref{tb:tinyImg} shows the results. One can see that in this experiment SNGAN is the best performer that achieves 57.74 inception score with SGD, and 86.30 with SVGD, 89.65 with LSVGD. In terms of classification accuracy on generated images, SNGAN achieves the highest 59.26\% with SGD, 76.77\% with SVGD and 78.37\% with LSVGD. LSVGD performs the best on all these GAN models, and on more than half of these GAN models LSVGD outperforms SGD and SVGD by at least 3 under inception score, and 5\% under classification accuracy. The last column of Table.~\ref{tb:tinyImg} also shows that LSVGD yields larger diversity compared with SGD and SVGD.

\subsection{Experiments on CelebA Dataset}

We also evaluate the performance of our method on a face attributes dataset --- CelebA \cite{liu2015faceattributes}, which contains more than 200K images in total. We sample 50k images as training set and another 10k images as validation set, and resize all these images to 64 $\times$ 64. On this dataset, we conduct an attribute-based image generation. We select 19 most active attributes (binary) from the entire dataset which are then used as side information input to generator. We also train an attribute classifier\footnote{We finetune a pretrained VGG-face network by replacing its outermost layer with attribute classifier, i.e., 19 binary classifiers corresponding to the 19 binary attributes.} to evaluate the quality of generated images. We report both the accuracy of attribute classification and the variance of generated images. The other settings are the same as Tiny-ImageNet.

Table.~\ref{tb:CelebAKL} shows the results. One can see that on most of the GAN models except DCGAN and WGAN-GP, LSVGD achieves higher accuracy than SGD and SVGD. We also can see that LSVGD raises diversity of generated image indicated by the magnitude of feature variance. Fig.~\ref{fig:celebImg} illustrates some generated examples which show that original training approach is a cause of mode collapse. SVGD helps to improve diversity but lowers the image quality. Fortunately, LSVGD maintains a good balance between image quality and diversity, and thus is an appropriate approach for training GANs.

\section{Conclusion}\label{sec_conclusion}
This paper proposes a stable particle-based method for variational inference --- Langevin SVGD. It is further incorporated into the generative adversarial net' framework to enhance the stability of GAN training and diversity of image generation. Its applicability is verified on a synthetic dataset and three popular benchmark datasets with promising results. This work provides a novel perspective of the relationship between GAN training and Bayesian inference, which therefore motivates Bayesian approaches to tackle the training stability issue of GAN.

\section{Appendix}
\subsection{Appendix A: Proof of Lemma~1}
This proof directly follows \cite{liu2016stein}. Since $T(\cdot)$ is set to be a bijective and differentiable function, therefore for any fixed $\epsilon$, ${T_\epsilon}(x) = x+ \epsilon + \delta \phi(x+ \epsilon)$ is also a bijective and differentiable function, then we have:
\begin{equation}
\mathrm{KL}(q_{[T_\epsilon]}||p) = \mathrm{KL}(q||p_{[T_\epsilon^{-1}]})
\end{equation}
where $p_{[T_\epsilon^{-1}]}$ denotes the distribution of $T_\epsilon^{-1}(x)$, then we have
\begin{equation}
\frac{\partial}{\partial\delta} \mathrm{KL}(q_{[T_\epsilon]}||p) = -E_{x\sim q}[\frac{\partial}{\partial\delta} \log p_{[T_\epsilon^{-1}]}(x) ]
\end{equation}
According to $p_{[T_\epsilon^{-1}]}(x) = p_{[T_\epsilon]}(x)|\mathrm{det} \nabla_x T_\epsilon(x)|$, then we have
\begin{equation}\label{eq_kld}
\begin{split}
\frac{\partial}{\partial\delta} \log p_{[T_\epsilon^{-1}]}(x) =& S_p(T_\epsilon(x))^T \frac{\partial}{\partial\delta} T_\epsilon(x)+ \\
&\mathrm{tr}\{(\nabla_x T_\epsilon(x))^{-1}\cdot \frac{\partial}{\partial\delta} \nabla_x T_\epsilon(x))\}
\end{split}
\end{equation}
Consider the limit as step size $\delta$ goes to $0$, we have
\begin{equation}\label{eq_Tapp}
\begin{split}
T_\epsilon(x) &= x+ \epsilon\\
\frac{\partial}{\partial\delta} T_\epsilon(x) &= \phi(x+ \epsilon)\\
\nabla_x T_\epsilon(x) &= I\\
\frac{\partial}{\partial\delta} \nabla_x T_\epsilon(x) &= \nabla_x \phi(x+ \epsilon)\\
\end{split}
\end{equation}
\begin{equation}
\begin{split}
\frac{\partial}{\partial\delta} \log p_{[T_\epsilon^{-1}]}(x) &= S_p(x+\epsilon)^T \phi(x+ \epsilon)+ \mathrm{tr}\{\nabla_x \phi(x+ \epsilon)\}\\
\end{split}
\end{equation}
Then, we have
\begin{equation}
\begin{split}
\frac{\partial}{\partial\delta} \mathrm{KL}\left(q_{[T_\epsilon]}||p\right)= -E_{x\sim q(x-\epsilon)}\mathrm{tr} \left[\mathcal{A}_{p}\phi(x)\right]
\end{split}
\end{equation}
Hence,
\begin{equation}\label{eq_res}
\begin{split}
E_{\epsilon\sim \mathcal{N}(0,\Sigma(x))} \frac{\partial}{\partial\delta} \mathrm{KL}\left(q_{[T_\epsilon]}||p\right)= -E_{x\sim q_\epsilon}\mathrm{tr} \left[\mathcal{A}_{p}\phi(x)\right]
\end{split}
\end{equation}
Eq.~\eqref{eq_res} holds because $q(x)$ can be approximated as:
\begin{equation}
q(x) \approx \frac{1}{n}\sum_{i=1}^n N(x_i,\sigma_0^2I)
\end{equation}
with $\sigma_0 \rightarrow 0$. Since $\epsilon$ and $-\epsilon$ have the same distribution, we have
\begin{equation}\label{eq_gmm}
\begin{split}
&E_{\epsilon(x)\sim \mathcal{N}(0,\Sigma(x)),~x\sim q(x-\epsilon)} = E_{\epsilon\sim \mathcal{N}(0,\Sigma(x)),~x\sim q(x+\epsilon)}\\
&= \frac{1}{n}\sum_{i=1}^n N(x_i,\Sigma(x_i)) = q_\epsilon(x)
\end{split}
\end{equation}
Combining Eq.\eqref{eq_res} and Eq.\eqref{eq_gmm}, we get the result of Theorem~1.
\subsection{Appendix B: Proof of Theorem~1}
In iteration $t$, a general updating rule for SGMCMC (eq.~(6) of \cite{Ma2015SGMCMC}) is:
\begin{equation}\label{eq_EM}
x' =x+\delta [(B(x)+Q(x))\nabla_{x}\log p(x)-\Gamma(x)]+ \epsilon(x)
\end{equation}
with
\begin{equation}
\Gamma^d(x)= \sum_{l=1}^D \frac{\partial}{\partial x^l}(B^{d,l}(x) + Q^{d,l}(x))
\end{equation}
In the case of LSVGD, $Q = 0$ and $B(x)=\sum_{i=1}^n 1\{x_i \in \Omega_x^R\} w_i k_{\gamma}(x_i,x) I_D$. Therefore,
\begin{equation}
\Gamma^d(x)=  \frac{\partial}{\partial x^d}B^{d,d}(x)
\end{equation}
Since for Gaussian kernel, we have
\begin{equation}
\nabla_{x_i}k_{\gamma}(x_i,x)= -\nabla_{x}k_{\gamma}(x_i,x)
\end{equation}
hence
\begin{equation}\label{eq_Gamma}
\begin{split}
\Gamma(x)&= \sum_{i=1}^n 1\{x_i \in \Omega_x^R\}\cdot w_i\cdot \nabla_{x}k_{\gamma}(x_i,x)\\
&= - \sum_{i=1}^n 1\{x_i \in \Omega_x^R\}\cdot w_i\cdot \nabla_{x_i}k_{\gamma}(x_i,x)
\end{split}
\end{equation}
Plugging Eq.~\eqref{eq_Gamma} and $B(x)$ back into Eq.~\eqref{eq_EM} recovers the update of LSVGD as Eq.~\eqref{eq_LSVGDfm}. Then, according to \cite{Ma2015SGMCMC}, by adjusting the covariance matrix of the injected noise as $\Sigma(x)=\delta(2B(x)-\delta \hat{V}(x))$, particles will eventually converge to the stationary distribution $p(x)$.

\subsection{Appendix C: Proof of Theorem~2}
We begin with the proof of Theorem 1. For notational convenience, denote $p(x+\epsilon)$ by $p_{\epsilon}(x)$, then
\begin{equation}
\begin{split}
E_{x\sim q_\epsilon}\mathrm{tr}[\mathcal{A}_p\phi(x)] &= E_{x\sim q,\epsilon\sim\mathcal{N}(\epsilon|0,\Sigma)}\mathrm{tr}[\mathcal{A}_{p_\epsilon}\phi(x+\epsilon)]
\end{split}
\end{equation}
where
\begin{equation}
\begin{split}
\mathrm{tr}[\mathcal{A}_{p_\epsilon}\phi(x+\epsilon)] = S_p(x+\epsilon)^T \phi(x+ \epsilon) + \mathrm{tr}[\nabla_x \phi(x + \epsilon)]
\end{split}
\end{equation}
As claimed in Theorem.~2, we can assume that for a sufficiently small $\epsilon$, all $\phi(x)$, $\nabla_x\phi(x)$ and $S_p(x)$ can be well approximated with the first order Taylor expansion:
\begin{equation}
\begin{split}
\phi(x+ \epsilon) &\approx \phi(x) + (\nabla_x\phi(x))^T\epsilon\\
\frac{\partial \phi(x+ \epsilon)}{\partial x_d} &\approx \frac{\partial \phi(x)}{\partial x_d} + \frac{\partial^2 \phi(x)}{\partial x_d^2}\epsilon_d\\
S_p(x+ \epsilon) &\approx S_p(x) + H_p(x)\epsilon
\end{split}
\end{equation}
where $H_p(x)$ is the Hessian matrix of $\log p(x)$ at $x$. Then
\begin{equation}
\begin{split}
E_\epsilon S_p(x+\epsilon)^T \phi(x+ \epsilon) \approx S_p(x)^T \phi(x) + E_\epsilon \epsilon^T \nabla_x\phi(x) H_p(x)\epsilon
\end{split}
\end{equation}
\begin{equation}
\begin{split}
E_\epsilon\mathrm{tr}[\nabla_x \phi(x + \epsilon)] &\approx E_\epsilon \sum_{d=1}^D \frac{\partial \phi(x+ \epsilon)}{\partial x_d}\\
&= \sum_{d=1}^D \frac{\partial \phi(x)}{\partial x_d}= \mathrm{tr}[\nabla_x \phi(x)]
\end{split}
\end{equation}
Therefore,
\begin{equation}
\begin{split}
\frac{\partial}{\partial\delta} \mathrm{KL}(q_{[T_\epsilon]}||p) &= -E_{x\sim q_\epsilon}\mathrm{tr}(\mathcal{A}_p\phi(x))\\
&\approx -\{E_{x\sim q}\mathrm{tr}(\mathcal{A}_p\phi(x)) - R(q,p)\}\\
&= \frac{\partial}{\partial\delta}\mathrm{KL}(q_{[T]}||p) + R(q,p)\\
\end{split}
\end{equation}
where $R(q,p)= -E_\epsilon \epsilon^T \nabla_x\phi(x) H_p(x)\epsilon$.


\subsection{Example of Gaussian mixture model}

The density of Gaussian mixture model can be rewritten as:
\begin{equation}
\begin{split}
p(x) &= \sum_{k=1}^K \pi_k \mathcal{N}(x| \mu_k, \sigma_k^2I)\\
 &= \sum_{z}\mathcal{N}(x| \mu_k, \sigma_k^2I)^{z_k}\pi_k^{z_k}
\end{split}
\end{equation}
where $z$ is the corresponding assignment variable of $x$ (i.e., $z_k = 1\{x \in \Omega_k\}$). Let $z$ use a $1$-of-$K$ representation, then the logarithm of $p$ reduces to,
\begin{equation}\label{eq_logGMM}
\log p(x) = \sum_{z}1\{z_k=1\} \cdot \log (\pi_k \mathcal{N}(x| \mu_k, \sigma_k^2I))
\end{equation}
then $H_p(x)= -\sum_k 1\{x \in \Omega_k\}\cdot \sigma_k^{-2}I$. For simplicity, let the weighting matrix $\nabla_x\phi(x)$ be a constant matrix, i.e., $\nabla_x\phi(x) = -C_{\phi}I$. Then we have,
\begin{equation}
|R(q, p)|\approx d\sigma^2 C_{\phi}\sum_k \sigma_k^{-2}E_{x\sim q}1\{x \in \Omega_k\}
\end{equation}

\bibliographystyle{IEEEtran}
\bibliography{refbibtex}
\end{document}